\documentclass[journal]{IEEEtran}
\ifCLASSINFOpdf
\else
\fi
\usepackage{orcidlink}
\usepackage{authblk}
\usepackage{colortbl}
\usepackage{microtype}
\usepackage{graphicx}
\usepackage{subfigure}
\usepackage{xcolor}
\usepackage{algorithm}
\usepackage{algpseudocode}
\usepackage{xcolor}
\usepackage{mdframed}
\usepackage{paralist}
\usepackage{amssymb}
\usepackage{booktabs}
\usepackage{multirow}
\usepackage{dsfont}
\usepackage{enumitem}
\usepackage{titlesec}
\usepackage{amsmath}
\usepackage{algpseudocode}
\usepackage{float} % Required for the [H] placement option
\usepackage{enumitem} % Required for the 'itemize' customization
\usepackage{mathtools}

\usepackage{tabularx} % for tabularx environment
\usepackage{array}    % for newcolumntype
\usepackage{hyperref}

\begin{document}
\title{Controllable Complex Human Motion Video Generation via Text-to-Skeleton Cascades}

\newcommand{\greenrow}{\rowcolor[HTML]{C6FFC6}}
\newcommand{\redrow}{\rowcolor[HTML]{ffafaf}}
\newcommand{\etal}{et~al.}
\newcommand{\argmaxA}{\operatorname*{argmax}}
\newcommand{\indep}{\perp\!\!\!\perp}
\newcommand{\MemberIEEE}{\textit{Member, IEEE}}
\newcommand{\SeniorMemberIEEE}{\textit{Senior Member, IEEE}}
\newcommand{\FellowIEEE}{\textit{Fellow, IEEE}}
\newcommand{\LifeFellowIEEE}{\textit{Life Fellow, IEEE}}

\author{Ashkan Taghipour\,\orcidlink{0000-0002-1950-5143},
        Morteza Ghahremani\,\orcidlink{0000-0001-6423-6475},
        Zinuo Li\,\orcidlink{0000-0003-0945-6621}, \MemberIEEE,

        Hamid Laga\,\orcidlink{0000-0002-4758-7510},
        Farid Boussaid\,\orcidlink{0000-0001-7250-7407}, \SeniorMemberIEEE,
        and Mohammed Bennamoun\,\orcidlink{0000-0002-6603-3257}, \SeniorMemberIEEE

\thanks{Ashkan Taghipour, Zinuo Li, and  Mohammed Bennamoun are with the Department of Computer Science and Software Engineering, The University of Western Australia, Australia. (Email: ashkan.taghipour@research.uwa.edu.au; Zinuo.li@research.uwa.edu.au, mohammed.bennamoun@uwa.edu.au)}
\thanks{Morteza Ghahremani is with the Munich Center for Machine Learning (MCML) and Technical University of Munich (TUM), Germany. (Email: morteza.ghahremani@tum.de)}
\thanks{Hamid Laga is with the School of Information Technology, Murdoch University, Australia.
(Email: h.laga@murdoch.edu.au)}
\thanks{Farid Boussaid is with the Department of Electrical, Electronics and Computer Engineering, The University of Western Australia, Australia.
(Email: farid.boussaid@uwa.edu.au).
 }
% \thanks{Manuscript received April 19, 2005; revised August 26, 2015.}
}

\markboth{}%IEEE Transactions on Image Processing
{Shell \MakeLowercase{\textit{et al.}}:  Demo of IEEEtran.cls for IEEE Journals}

\maketitle

\begin{abstract}
Generating videos of complex human motions—such as flips, cartwheels, and martial arts—remains challenging for current video diffusion models. Text-only conditioning is temporally ambiguous for fine-grained motion control, while explicit pose-based controls, though effective, require users to supply complete skeleton sequences that are costly to generate for long, dynamic actions. We propose a two-stage cascaded framework that addresses both limitations. First, an autoregressive text-to-skeleton model generates 2D pose sequences from natural language descriptions, predicting each joint conditioned on previously generated poses to capture long-range temporal dependencies and inter-joint coordination in complex motions. Second, a pose-conditioned video diffusion model synthesizes videos from a reference image and the generated skeleton sequence, employing DINO-ALF (Adaptive Layer Fusion), a multi-level reference encoder that preserves appearance and clothing details under large pose changes and self-occlusions. To address the lack of publicly available datasets for complex human motion video generation, we introduce a Blender-based synthetic dataset of 2,000 videos featuring diverse characters performing acrobatic and stunt-like motions, providing full control over appearance, motion, and environment. This dataset fills a critical gap, as existing benchmarks severely under-represent acrobatic and stunt-like motions, while also avoiding the copyright and privacy concerns of web-collected data. Experiments on our proposed synthetic dataset and the Motion-X Fitness benchmark demonstrated that our text-to-skeleton model outperformed prior methods on FID, R-precision, and motion diversity, while our pose-to-video model achieved the best results among all compared methods on VBench metrics for temporal consistency, motion smoothness, and subject preservation. Additional results are available at the \href{https://ashkantaghipour.github.io/kangaroo/}{Project Page}.
\end{abstract}


% \fi
\IEEEpeerreviewmaketitle

\section{Introduction}
\IEEEPARstart{D}{iffusion} models have emerged as the dominant generative framework in computer vision~\cite{TIP_1,TIP_2,TIP_3, taghipur_b2b, human_video_survey}. Extended to video, they have enabled remarkable progress in text-to-video (T2V)~\cite{ltx, hunyuanvideo} and text-and-image-to-video (TI2V) generation~\cite{cogvideox, svd}. In TI2V, a reference image defines the subject's appearance while text describes the desired motion, enabling the synthesis of photorealistic content with realistic motions~\cite{skyreels, taghipour_video}. Despite these advances, the controllable generation of complex human motions, such as flips, cartwheels, acrobatics, and martial arts, remains an open challenge~\cite{animateanyone, magicanimate}. In this work, we refer to complex human motion as non-repetitive, highly dynamic actions that involve large pose changes and frequent self-occlusions.

Current large-scale video diffusion models, both open-source~\cite{wan, hunyuanvideo, cogvideox} and proprietary~\cite{sora, kling}, struggle with such motions, often producing implausible limb trajectories, temporal inconsistency in body shape or clothing, and appearance drift~\cite{hypermotion}. These failures limit applications in sports content creation~\cite{echomotion}, virtual coaching~\cite{one_to_all}, stunt pre-visualization~\cite{steadydancer, wananimate}, and avatar animation~\cite{4d_animation}.

A core challenge in TI2V generation of complex motion is that text alone provides insufficient control~\cite{vedaldi}. Descriptions like ``a person performing a backflip'' are semantically clear but temporally ambiguous—they do not specify frame-wise joint trajectories or sub-motion timing. To address this, recent works condition on explicit motion signals such as 2D skeletons or depth maps, significantly improving controllability~\cite{steadydancer, champ, animateanyone2}. However, these methods require users to supply complete pose sequences, which is impractical for complex actions: generating high-quality 2D poses is time-consuming and demands specialized tools~\cite{dreamoving, dreampose}. As a result, practitioners are constrained to small libraries of template motions, limiting expressiveness and scalability.

Even when pose sequences are available, preserving the reference appearance under complex motion remains challenging. Existing pose-conditioned methods encode the reference image using CLIP-based embeddings injected via cross-attention~\cite{animateanyone, magicanimate, champ}. CLIP-based conditioning works for structured motions with moderate pose changes, but struggles under large deformations, rapid transitions, and self-occlusions that are common in complex actions. CLIP produces global, semantic representations that lack fine-grained spatial detail~\cite{clearclip, cloc}, making it difficult to reconstruct local appearance cues such as clothing textures and body parts under viewpoint changes or occlusion~\cite{clip2dino}. As a result, these models exhibit appearance inconsistency, texture blurring, and loss of body part details during dynamic motion~\cite{consisid, mofe}.

These observations motivate a two-stage cascaded framework that addresses both challenges. In the first stage, an autoregressive text-to-skeleton model translates natural language into 2D pose sequences, capturing long-range temporal dependencies and inter-joint coordination for complex, non-repetitive actions without requiring manual pose generation. In the second stage, a pose-conditioned video diffusion model synthesizes videos from a reference image and the generated skeleton sequence, using DINO-ALF (Adaptive Layer Fusion) to preserve appearance under large deformations and self-occlusions by adaptively aggregating spatially localized patch descriptors across multiple DINO layers. To train and evaluate on complex motions, we also construct a Blender-based synthetic dataset of 2,000 videos featuring acrobatic and stunt-like actions, filling a gap left by existing benchmarks that focus on regular, repetitive activities.

Building upon the insight that controllable generation of complex human motion requires explicitly decoupling motion planning from appearance synthesis, our main contributions are as follows:
\begin{itemize}
    \item \textbf{Motion planning via autoregressive text-to-skeleton generation.}
    We propose an autoregressive text-to-skeleton model that translates natural language descriptions into joint-level 2D pose sequences, explicitly modeling long-range temporal dependencies and inter-joint coordination. Unlike prior text-to-pose methods that focus on pose realism alone, our formulation produces structured motion plans that serve as an explicit and editable control signal for complex, non-repetitive actions, eliminating the need for manual pose generation.

    \item \textbf{Deformation-aware pose-conditioned video diffusion with DINO-ALF.}
    We introduce DINO-ALF (Adaptive Layer Fusion), a deformation-aware appearance conditioning mechanism for pose-conditioned video diffusion. DINO-ALF leverages spatially localized patch descriptors and adaptively aggregates complementary features across multiple DINOv3 layers to maintain appearance correspondence under large pose deformations and self-occlusions. This design enables robust preservation of identity and clothing details.

    \item \textbf{A synthetic dataset for complex human motion.}
    We construct and release a Blender-based synthetic dataset of 2,000 complex-motion videos specifically targeting acrobatic and stunt-like actions underrepresented in existing benchmarks, while also avoiding the copyright and privacy concerns associated with web-collected data.
\end{itemize}

\section{Related Work}
\label{sec:related}

\subsection{Pose-Conditioned Human Video Generation}

Pose-conditioned video generation addresses the limitations of text-only control by introducing explicit structural signals such as 2D skeletons, depth maps, or parametric body models to guide human animation~\cite{animateanyone, magicanimate, champ}. Two core challenges define this paradigm: how to inject pose control into the video denoiser, and how to preserve the reference appearance throughout the generated sequence.

\noindent\textbf{Pose control injection.}
Different architectural choices govern how structural controls are fused with the video denoiser. ControlNet-style adapters~\cite{animateanyone, controlnext} inject pose guidance through trainable copy branches that add residuals to frozen backbone layers. DreamPose~\cite{dreampose} and DreaMoving~\cite{dreamoving} fine-tune such adapters for fashion and character animation respectively, while Follow-Your-Pose~\cite{followyourpose} enables pose-guided generation using pose-free training videos. DiT-based architectures enable alternative injection patterns: VACE~\cite{vace} renders poses as RGB control videos, encodes them into spatiotemporally aligned context tokens, and injects these via dedicated Context Blocks; Human4DiT~\cite{human4dit} extends pose-conditioned generation to free-viewpoint synthesis with 4D diffusion transformers; UniAnimate~\cite{unianimate} adapts feature-encoder conditioning to DiT backbones. HumanVid~\cite{humanvid} disentangles camera and body motion for camera-controllable animation, while SteadyDancer~\cite{steadydancer} and Wan-Animate~\cite{wananimate} refine temporal coherence and first-frame preservation.

\noindent\textbf{Reference appearance encoding.}
Preserving the reference appearance while following a driving pose is equally critical. Early methods encode the reference image using CLIP embeddings injected via cross-attention~\cite{animateanyone, dreampose}. However, CLIP's global semantic representations lack fine-grained spatial details, causing appearance drift under large deformations~\cite{clearclip, cloc}. To address this, MagicAnimate~\cite{magicanimate} introduces ReferenceNet, a framework that duplicates the denoiser's spatial layers to extract dense appearance features. Champ~\cite{champ} augments pose guidance with 3D parametric cues from SMPL shape and normal maps. DisCo~\cite{disco} and MagicPose~\cite{magicpose} disentangle motion from appearance through separate control pathways or multi-stage training. MimicMotion~\cite{mimicmotion} strengthens guidance in high-frequency failure regions such as hands and faces, while HyperMotion~\cite{hypermotion} proposes a spatial low-frequency enhanced RoPE design for fast pose changes. StableAnimator~\cite{stableanimator} and Animate-X~\cite{animate_x} refine human-specific fusion of pose and reference features, AnimateAnyone 2~\cite{animateanyone2} incorporates environment affordance, SCAIL~\cite{scail} leverages 3D-consistent pose representations, and EchoMotion~\cite{echomotion} unifies video and motion generation via dual-modality diffusion.

While these methods have advanced controllable human animation, most are validated on relatively regular dynamics such as dance sequences. Generating complex, non-repetitive motions remains challenging: most pipelines assume user-provided pose sequences, which are impractical to generate for dynamic actions~\cite{dreampose, dreamoving}. Also, existing reference encodings struggle under large deformations and self-occlusions~\cite{consisid, mofe}.

\begin{figure*}[t]
    \centering
    \includegraphics[width=1\textwidth]{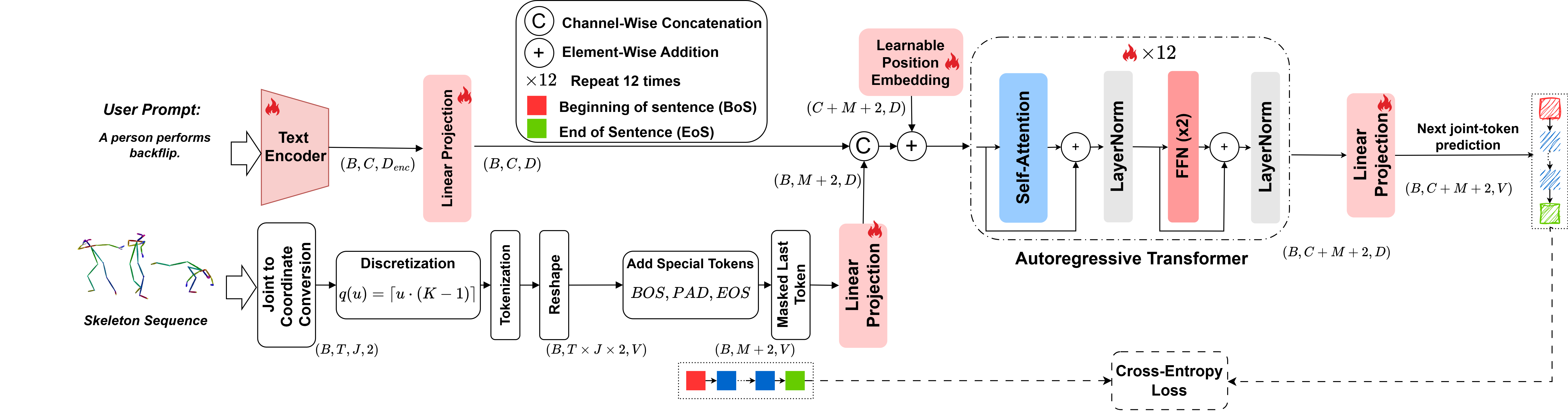}
    \caption{Overview of the text-to-skeleton generation architecture for training. A text prompt is encoded and prepended as a conditioning prefix to the pose token sequence. The autoregressive Transformer predicts each joint token conditioned on all previously generated tokens and the text description.}
    \vspace{-6pt}
    \label{fig:train_motion_diagram}
\end{figure*}

\subsection{Text-to-Skeleton as Motion Control for Video}
\noindent A practical limitation of pose-guided human video generation is that it assumes users can provide a full future pose sequence \cite{scail, steadydancer, wananimate}, which is costly for long and complex actions. To reduce this burden while retaining an explicit and editable control interface, recent work generates intermediate motion controls from language, most commonly 2D skeleton sequences that can directly drive pose-conditioned video generators \cite{humandreamer, signllm, signmix, signidd}. This choice is well aligned with image-space video diffusion, where 2D pose provides a view-aligned and easily editable structural trace without introducing additional camera or depth ambiguities \cite{holistic_motion_2d, humandreamer}.

HumanDreamer~\cite{humandreamer} follows this paradigm by generating 2D pose sequences from text and then driving pose-to-video generators, proposing a DiT-based pose generator with an additional latent alignment objective to better match language and motion. Holistic-Motion2D~\cite{holistic_motion_2d} and its extension Tender scale text-to-2D motion learning with whole-body keypoints and introduce part-aware and confidence-aware modeling to better handle noisy detections and occlusions in 2D pose trajectories. Motion-2-to-3~\cite{motion_2d_to_3d} models 2D motion in a disentangled form and extends generation to multi-view settings for view-consistent motion. From a complementary perspective, Mimic2DM~\cite{Mimic2DM} learns control from in-the-wild 2D keypoint trajectories and also employs an autoregressive transformer to generate 2D reference motions inside a hierarchical control framework. MoSA~\cite{mosa} generates human keypoints from text in a separate structure stage and uses the projected 2D skeletons as guidance for a DiT-based video generator. Human-Motion2D Generation~\cite{2d_skelet} proposes a diffusion model that generates 2D skeleton sequences from text (optionally conditioned on an initial motion frame), and further improves realism and alignment via a reinforcement learning (RL)-style fine-tuning stage, showing that the generated skeletons can serve as explicit controls for skeleton-guided video generation.

Text-to-skeleton generation also enables sign language video synthesis, where skeleton sequences serve as intermediate control for rendering into human appearance~\cite{survey_sign}. Mixed SIGNals~\cite{signmix} generates sign pose sequences by blending motion primitives, Sign-IDD~\cite{signidd} enforces skeletal consistency via bone direction and length constraints, and SignLLM~\cite{signllm} outputs skeletal representations that drive pose-to-video synthesis.

Despite this progress, existing text-to-2D pose models are typically developed for general motions or for specific downstream settings, and their ability to generate highly dynamic, \emph{non-repetitive} pose trajectories involving rapid transitions and self-occlusions remains less explored. In particular, highly dynamic actions with rapid transitions and self-occlusions place stronger demands on temporal coherence and inter-joint coordination than the relatively regular motions often emphasized in prior benchmarks.

\section{Proposed Method}

The proposed method follows a two-stage cascaded framework designed to generate explicit motion control from text while preserving appearance under complex movement. Several design choices are motivated by the specific demands of complex and non-repetitive actions. We operate in 2D pose space rather than 3D representations, since 2D skeletons are directly aligned with the image plane of the video diffusion model, avoiding the camera-projection ambiguities that arise with 3D representations. We adopt an autoregressive factorization over discretized joint coordinates because complex motions exhibit strong sequential dependencies—each joint's position depends on the preceding trajectory and on the configuration of other joints—and discrete tokens allow us to model this distribution with a standard next-token objective while enabling controllable sampling strategies. For appearance conditioning, we extract features from multiple DINO layers rather than relying on a single CLIP embedding, because earlier DINO layers capture texture-rich local details while later layers encode more view-invariant semantics; adaptively fusing them provides the spatially localized cues needed to preserve identity under large deformations and self-occlusions. The first stage, Text-to-Skeleton Generation (Sec.~\ref{sec:text2pose}), translates natural language into 2D pose sequences, and the second stage, Pose-Conditioned Video Generation (Sec.~\ref{sec:pose2video}), synthesizes video frames from a reference image and the generated skeleton using DINO-ALF appearance conditioning.

\begin{figure*}[t]
    \centering
    \includegraphics[width=.84\textwidth]{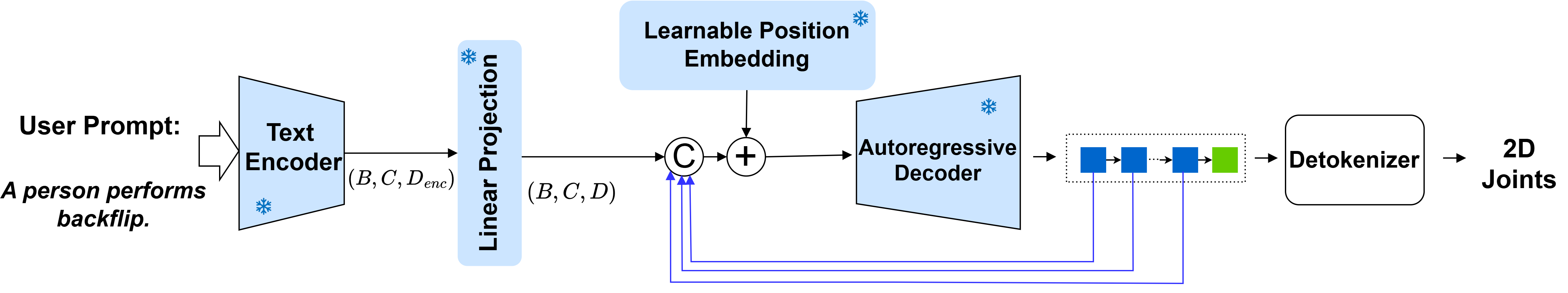}
    \caption{Overview of the text-to-skeleton generation architecture for inference.}
    \vspace{-6pt}
    \label{fig:inference_motion_diagram}
\end{figure*}

\subsection{Text-to-Skeleton Generation}
\label{sec:text2pose}

Our first stage maps a natural language motion description into a sequence of 2D skeleton keypoints (Fig.~\ref{fig:train_motion_diagram}). We represent a motion as a sequence of tokens, where each token corresponds to the 2D coordinates of a single joint at a particular time step. A Transformer-based autoregressive model~\cite{gendop} then predicts each joint token conditioned on all previously generated joints and the input text. This factorization is particularly suitable for complex human motion, as it exposes strong temporal and inter-joint dependencies to the model, while the fixed joint-ordering and bounded discrete vocabulary naturally encourage structurally valid pose sequences.

\subsubsection{Pose Representation and Tokenization}

Let a motion clip be represented by $T$ frames and $J$ joints per frame, $\mathbf{P}\in\mathbb{R}^{T\times J\times 2}$,
where each joint is parameterized by normalized image-plane coordinates
$(x_{t,j}, y_{t,j}) \in [0,1]^2$ for frame $t \in \{1,\dots,T\}$ and joint
index $j \in \{1,\dots,J\}$. In practice, we normalize pixel coordinates by the frame width and height so that all joints lie in $[0,1]^2$ before discretization. For notational brevity, we henceforth denote the coordinate' number ${T \times J \times 2}$ as $M$.

\noindent \textbf{Discrete coordinate tokens.}
To model 2D pose sequences with an autoregressive Transformer, we first convert the continuous pose tensor into discrete token IDs. Given a normalized coordinate $u \in [0,1]$, we discretize it into $K$ bins via
\begin{equation}
\label{eq:quantize}
q(u) = \lceil u \cdot (K-1)\rceil,
\qquad q(u)\in\{0,\dots,K-1\}.
\end{equation}
We reserve the first four token IDs for special symbols: $\texttt{PAD}=0$, $\texttt{BOS}=1$, $\texttt{EOS}=2$, and a reserved symbol. All body coordinate tokens are shifted upward by an offset $o=4$:
\begin{equation}
\label{eq:shift}
s(u)=q(u)+o,\qquad s(u)\in\{4,\dots,K+3\}
\end{equation}
yielding a total vocabulary size of $V = o + K$. This follows standard practice in discrete trajectory modeling, where auxiliary tokens occupy a reserved range that does not collide with data tokens~\cite{gendop}. With each coordinate now mapped to a discrete token, we next describe how to arrange these tokens into a sequence for autoregressive modeling.

\noindent \textbf{One-dimensional token stream.}
We serialize the 2D pose tensor into a one-dimensional token stream $\mathbf{z} = [z_1,\dots,z_M]$ of length $M = 2JT$, where each joint contributes two consecutive tokens for its $x$ and $y$ coordinates. We adopt a frame-major, joint-minor ordering so that the model completes an entire body pose before advancing in time, ensuring spatial coherence among all joints within each frame: for each frame, joints appear in a fixed skeleton order, each contributing consecutive $(x,y)$ tokens. The resulting sequence is:
\begin{multline}
\label{eq:streamdef}
\mathbf{z} = \bigl[\,s(x_{1,1}), s(y_{1,1}), \ldots, s(x_{1,J}), s(y_{1,J}),\\
\ldots\,,\, s(x_{T,J}), s(y_{T,J})\,\bigr].
\end{multline}
For autoregressive training, we prepend a \texttt{BOS} token and append an \texttt{EOS} token to form the full sequence:
\begin{equation}
\label{eq:full_seq}
\mathbf{z}^{\text{full}} = [\texttt{BOS}, z_1, \ldots, z_M, \texttt{EOS}].
\end{equation}

\subsubsection{Text conditioning}

Given a natural-language motion description $c$ (e.g., ``a person performs a flying knee punch''), we encode it using the frozen CLIP text encoder from Stable Diffusion v2~\cite{sdv2}. Tokenizing $c$ and passing it through the encoder produces a sequence of $C$ contextual embeddings
\[
\mathbf{h}^{\text{text}} = [h^{\text{text}}_1,\dots,h^{\text{text}}_C]
\in \mathbb{R}^{C\times D_{\text{enc}}}.
\]
These embeddings are projected to the Transformer's hidden dimension and normalized:
\begin{equation}
\label{eq:textproj}
\mathbf{e}^{\text{cond}} = \operatorname{LN}\big(\mathbf{h}^{\text{text}} W_{\text{cond}}\big),
\end{equation}
where $W_{\text{cond}} \in \mathbb{R}^{D_{\text{enc}}\times D}$ is a learned linear projection and $\operatorname{LN}(\cdot)$ denotes layer normalization. The resulting sequence $\mathbf{e}^{\text{cond}} \in \mathbb{R}^{C \times D}$ is prepended to the pose token sequence and remains visible to every subsequent token through causal self-attention, ensuring that text conditioning persists throughout the entire generation process.

\subsubsection{Autoregressive Decoder}

To process both text and pose tokens in a single autoregressive model, we embed each token in $\mathbf{z}^{\text{full}}$ using a learned embedding table $E_{\text{pose}} \in \mathbb{R}^{V \times D}$, where $V$ is the vocabulary size defined in Eq.~\eqref{eq:shift} and $D$ is the Transformer hidden dimension. This produces a sequence of pose embeddings $\mathbf{E}^{\text{pose}} \in \mathbb{R}^{(M+2) \times D}$, where $M+2$ accounts for the $M$ body-coordinate tokens plus the \texttt{BOS} and \texttt{EOS} tokens.

We concatenate the text-conditioning sequence and the pose embeddings along the sequence dimension:
\begin{equation}
\label{eq:full_input}
\mathbf{H}^{(0)} = [\mathbf{e}^{\text{cond}}, \mathbf{E}^{\text{pose}}]
\in \mathbb{R}^{(C+M+2)\times D}.
\end{equation}
We add learned positional embeddings to $\mathbf{H}^{(0)}$ and feed the result into a stack of $L$ decoder-only Transformer blocks with causal self-attention~\cite{gendop}, producing hidden states $\mathbf{H}^{(L)} \in \mathbb{R}^{(C+M+2)\times D}$. The first $C$ positions correspond to the text prefix, and the remaining $M+2$ positions correspond to \texttt{BOS}, the body tokens, and \texttt{EOS}.

A linear output head maps the final hidden states to logits over the vocabulary:
\begin{equation}
\label{eq:lm_head}
\mathbf{O} = \mathbf{H}^{(L)} W_{\text{lm}}^\top
\in \mathbb{R}^{(C+M+2)\times V},
\end{equation}
where $W_{\text{lm}} \in \mathbb{R}^{V\times D}$. Applying a softmax over the vocabulary dimension yields the text-conditioned autoregressive distribution:
\begin{equation}
\label{eq:ar}
p_\theta(\mathbf{z}^{\text{full}} \mid c)
= \prod_{i=1}^{M+1} p_\theta(z_i \mid \mathbf{e}^{\text{cond}}, z_{<i}),
\end{equation}
where $z_1, \ldots, z_M$ are the body-coordinate tokens, $z_{M+1} \triangleq \texttt{EOS}$, and $z_{<1}$ consists only of the \texttt{BOS} token. The text embeddings $\mathbf{e}^{\text{cond}}$ act as a fixed conditioning prefix throughout generation.

% ------------------------------------------------------------------------

\begin{figure*}[t]
    \centering
    \includegraphics[width=0.95\textwidth]{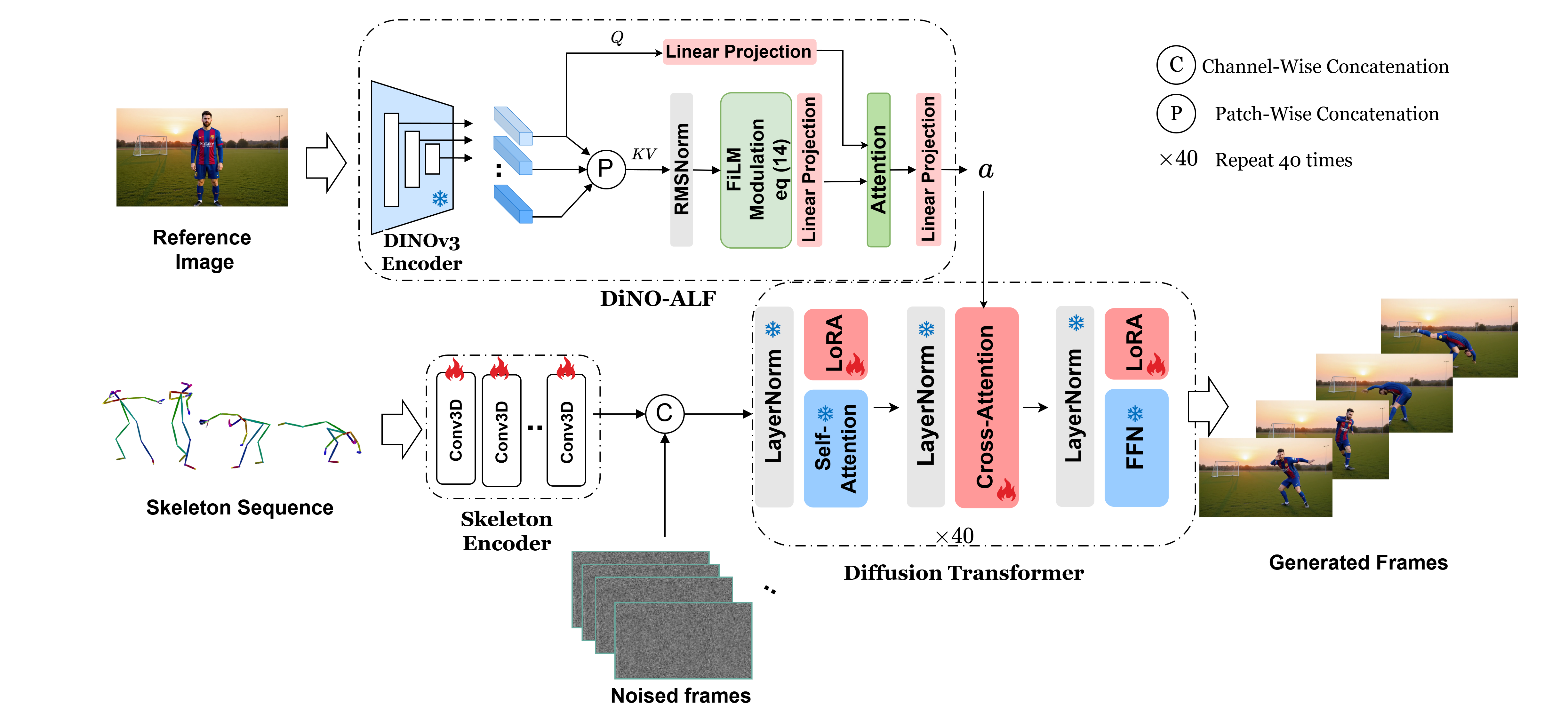}
    \caption{Overview of the pose-conditioned video generation architecture. A reference image is encoded via DINO-ALF to produce appearance tokens, while the skeleton sequence is rasterized and encoded by a 3D CNN into spatiotemporally aligned motion tokens. Both conditioning streams are injected into the DiT denoiser to synthesize the output video.}
    \vspace{-6pt}
    \label{fig:video_diagram}
\end{figure*}

\subsubsection{Training objective}

During training, we use teacher forcing, which conditions on ground-truth previous tokens rather than model predictions. We supervise only the body tokens and the final \texttt{EOS} token; the text prefix and \texttt{BOS} token serve as context and do not contribute to the loss.

The training objective is the standard next-token cross-entropy:
\begin{equation}
\label{eq:loss}
\mathcal{L}_{\text{text2pose}}
= -\sum_{i=1}^{M+1}
\log p_\theta\!\left(z_i \mid \mathbf{e}^{\text{cond}}, z_{<i}\right),
\end{equation}
where $z_1, \ldots, z_M$ are the body-coordinate tokens and $z_{M+1} = \texttt{EOS}$.

\subsubsection{Inference}
The inference process is illustrated in Fig.~\ref{fig:inference_motion_diagram}, where the model predicts the next token autoregressively until the \texttt{EOS} token is generated.

% ------------------------------------------------------------------------
\subsection{Pose-Conditioned Video Generation}
\label{sec:pose2video}

Our second stage synthesizes a video from two inputs: a reference image $I_{\mathrm{ref}}$ specifying \emph{who} should appear, and a skeleton sequence $\hat{\mathbf{P}}\in\mathbb{R}^{T\times J\times 2}$ specifying \emph{how} the body should move, where $\hat{\mathbf{P}}$ is generated by the text-to-skeleton model described in Section~\ref{sec:text2pose}. The output is a video in which the reference character performs the desired complex motion (Fig.~\ref{fig:video_diagram}).

Since $\hat{\mathbf{P}}$ is predicted rather than ground-truth, errors from the skeleton generation stage can propagate into video synthesis. To improve robustness, we apply stochastic augmentations to ground-truth skeletons during training, each mimicking a typical failure mode of the generation stage: (i)~\emph{joint jitter}---Gaussian noise ($\sigma{=}3$\,pixels) per coordinate; (ii)~\emph{joint dropout}---each joint zeroed with probability $p_j{=}0.05$; and (iii)~\emph{temporal shift}---positions displaced by ${\pm}1$ frame.

Following prior pose-guided human animation works~\cite{steadydancer,scail,wananimate}, we render $\hat{\mathbf{P}}$ into a per-frame 2D pose-control image by rasterizing joints and their skeletal connections, producing an image sequence spatially aligned with the video frames. We build on the pretrained Wan2.1 TI2V diffusion backbone~\cite{wan} and adapt it to pose-driven generation. Since motion is now explicitly specified by the skeleton sequence rather than text, we replace the text cross-attention pathway with pose-based conditioning.

Our design introduces three key components:
\begin{inparaenum}[(i)]
\item DINO-ALF, a multi-level DINOv3~\cite{simeoni2025dinov3} appearance encoder that extracts spatially localized patch features to preserve clothing, textures, and local body details under large pose changes;
\item a modified conditioning interface that replaces the CLIP-based reference cross-attention with a fully trainable DINO-ALF cross-attention, combined with LoRA adapters on selected DiT self-attention and MLP layers; and
\item a spatiotemporally aligned motion encoder that maps the rendered pose sequence into context tokens for explicit motion guidance.
\end{inparaenum}

% ------------------------------------------------------------------------
\subsubsection{Latent video diffusion backbone}
Let $\mathbf{v}\in\mathbb{R}^{3\times T\times H\times W}$ denote a training video clip with $T$ frames.
A pretrained spatiotemporal VAE encodes $\mathbf{v}$ into latent tokens $\mathbf{x}_0\in\mathbb{R}^{C_v\times T'\times H'\times W'}$, where $C_v$ is the latent channel dimension and $(T', H', W')$ denote the temporally and spatially downsampled resolution.
We follow standard diffusion training: sample a timestep $\tau$ and noise $\boldsymbol{\epsilon}\sim\mathcal{N}(\mathbf{0},\mathbf{I})$, and construct noisy latents
\begin{equation}
\label{eq:diff_forward}
\mathbf{x}_\tau=\alpha_\tau\,\mathbf{x}_0+\sigma_\tau\,\boldsymbol{\epsilon}.
\end{equation}
A DiT denoiser $\epsilon_{\theta}(\cdot)$ predicts the noise (or velocity, depending on the scheduler) conditioned on appearance and motion controls:
\begin{equation}
\label{eq:diff_denoise}
\hat{\boldsymbol{\epsilon}}=\epsilon_{\theta}\!\left(\mathbf{x}_\tau, \tau \mid I_{\mathrm{ref}}, \hat{\mathbf{P}}\right),
\end{equation}
and we minimize a weighted MSE objective
\begin{equation}
\label{eq:diff_loss}
\mathcal{L}_{\mathrm{vid}}=\lambda(\tau)\,\big\lVert \hat{\boldsymbol{\epsilon}}-\boldsymbol{\epsilon} \big\rVert_2^2,
\end{equation}
where $\lambda(\tau)$ is the scheduler-dependent loss weight.

% ------------------------------------------------------------------------
\subsubsection{DINO-ALF: Adaptive Layer Fusion for Appearance Encoding}
The pretrained backbone conditions generation on a reference image and text by concatenating CLIP image/text tokens and injecting them into the DiT blocks via cross-attention.
However, under complex motions with large deformations and self-occlusions, this global CLIP-style conditioning is often insufficient to preserve fine-grained appearance cues (e.g., clothing textures and local body parts), leading to noticeable drift~\cite{clearclip, cloc, clip2dino}.
To provide stronger, spatially localized reference cues, we extract and adaptively \emph{fuse multi-layer} patch descriptors from a frozen DINOv3 encoder and inject them as appearance tokens for conditioning.

\noindent\textbf{Multi-layer extraction.}
Let $\{\mathbf{h}^{(\ell)}\}_{\ell=1}^{L_D}$ be the hidden states from all $L_D{=}12$ DINO Transformer layers (patch size $16{\times}16$), and let
$\mathbf{p}^{(\ell)}\in\mathbb{R}^{N\times d_D}$ denote the corresponding \emph{patch tokens} (CLS/register tokens removed), where $N$ is the number of patches and $d_D$ is the embedding size of the encoder.

\begin{figure}[t]
    \centering
    \includegraphics[width=\columnwidth]{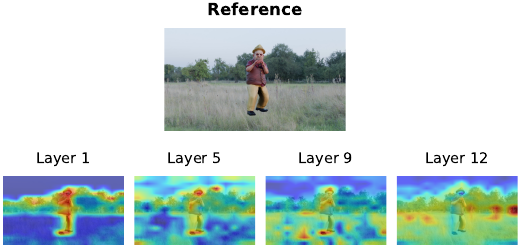}
    \caption{Patch-feature magnitude maps ($\ell_2$-norm) across DINOv3 layers. Earlier layers exhibit high activation on the subject and texture-rich regions, while later layers show more uniform magnitudes. This motivates using an early layer as the query for adaptive layer fusion.}
    \vspace{-6pt}
    \label{fig:dino_features}
\end{figure}

\begin{figure}[t]
    \centering
    \includegraphics[width=\columnwidth]{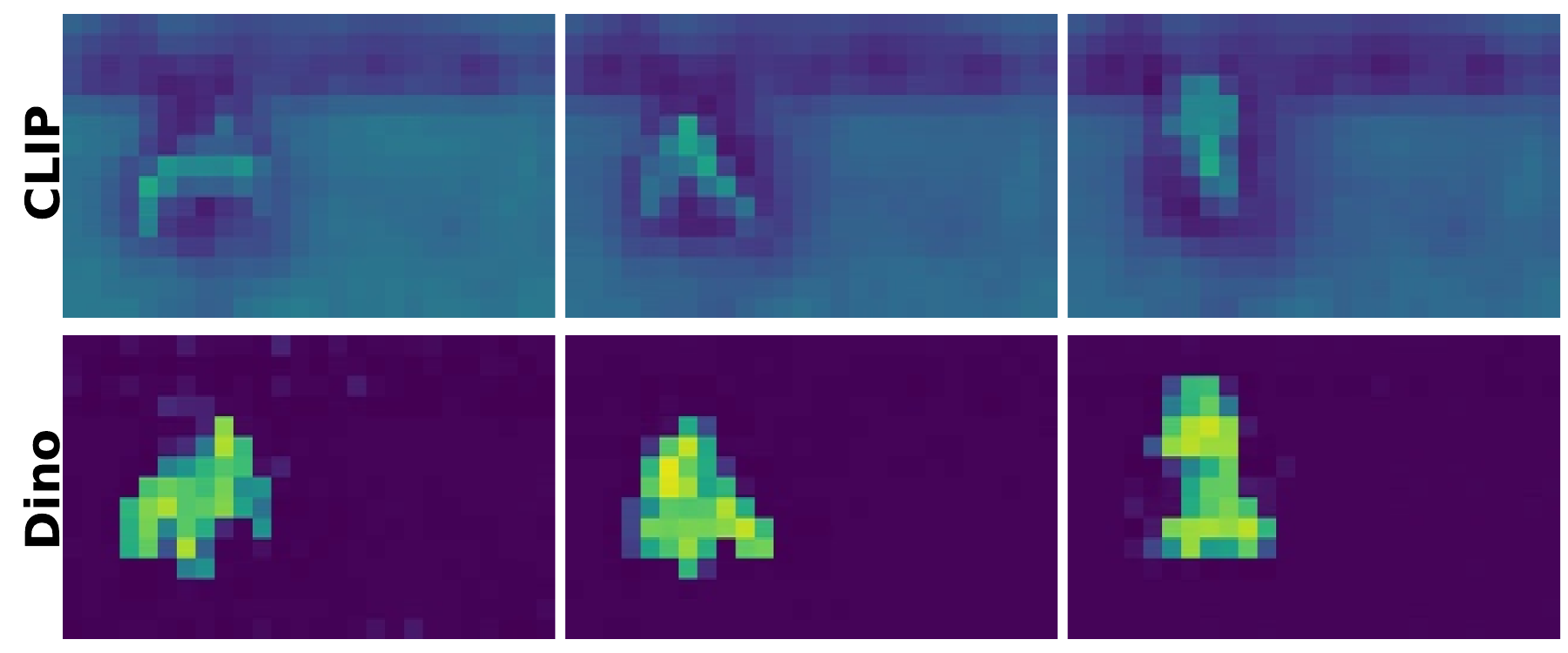}
    \caption{Cross-attention maps for CLIP (top) vs.\ DINO-ALF (bottom) on a backflip sequence. DINO-ALF attends more precisely to the moving subject, while CLIP attention is scattered.}
    \vspace{-6pt}
    \label{fig:cross_attention}
\end{figure}

\noindent\textbf{Adaptive layer fusion.}
Different DINO layers capture complementary cues: as shown in Fig.~\ref{fig:dino_features}, earlier layers exhibit high activation on texture-rich regions while later layers produce more uniform responses, motivating our use of an early layer as the query for adaptive fusion.
Instead of choosing a single layer, we learn to aggregate them per-patch via a cross-attention module $A_{\gamma}$. We first stack the layer-wise patch tokens:
\begin{equation}
\label{eq:dino_stack}
\mathbf{P}=[\mathbf{p}^{(1)},\dots,\mathbf{p}^{(L_D)}]\in\mathbb{R}^{N\times L_D\times d_D}.
\end{equation}
To stabilize training and let the model emphasize or suppress individual layers, we apply a FiLM-style modulation:
\begin{equation}
\label{eq:film}
\mathbf{P}'_{:,\,\ell,:}=\mathrm{Norm}(\mathbf{P}_{:,\,\ell,:})\odot(1+\boldsymbol{\beta}_{\ell})+\boldsymbol{\delta}_{\ell},
\end{equation}
with learnable $(\boldsymbol{\beta}_{\ell},\boldsymbol{\delta}_{\ell})$ for each layer $\ell$. Cross-attention then aggregates the modulated features:
\begin{equation}
\label{eq:dino_agg}
\tilde{\mathbf{p}}=A_{\gamma}(\mathbf{q},\mathbf{P}')\in\mathbb{R}^{N\times d_D},
\end{equation}
where $\mathbf{q}\in\mathbb{R}^{N\times d_D}$ is a query derived from the first layer ($\ell{=}1$).
Finally, we project the aggregated tokens into the DiT hidden dimension:
\begin{equation}
\label{eq:dino_proj}
\mathbf{a}=p_{\eta}(\tilde{\mathbf{p}})\in\mathbb{R}^{N\times d},
\end{equation}
where $p_{\eta}$ is a MLP with LayerNorm.
The DINO encoder remains frozen; only $A_{\gamma}$ and $p_{\eta}$ are trained. As shown in Fig.~\ref{fig:cross_attention}, the resulting DINO-ALF cross-attention remains concentrated on the human subject, whereas CLIP cross-attention is scattered across the image, enabling better alignment between motion and appearance.

% ------------------------------------------------------------------------

\subsubsection{Replacing native reference cross-attention with DINO cross-attention}
Wan2.1 injects conditioning through two cross-attention pathways per DiT block: one attending to CLIP text tokens to steer motion and scene evolution, and one attending to CLIP image tokens for appearance. In our pose-driven setting, motion is explicitly specified by the skeleton sequence, so we disable both CLIP-based pathways and replace them with a single DINO-conditioned cross-attention that attends to the DINO-ALF appearance tokens $\mathbf{a}$, providing spatially localized appearance cues that better preserve clothing and fine details under large pose changes.
% ------------------------------------------------------------------------
\subsubsection{Pose control as spatiotemporally aligned motion tokens}
Directly injecting raw 2D keypoints into a high-capacity video denoiser is ineffective unless the conditioning is \emph{aligned} with the latent spatiotemporal grid.
We therefore rasterize the skeleton sequence into an RGB pose-control video
$\mathbf{s}\in\mathbb{R}^{3\times T\times H\times W}$ (e.g., stick figures / keypoint lines) following the rendering setup of~\cite{xu2023ViTPose++}, and encode it with a 3D CNN motion encoder
$g_{\phi}$:
\begin{equation}
\label{eq:pose_tokens}
\mathbf{m}=g_{\phi}(\mathbf{s}), \qquad \mathbf{m}\in\mathbb{R}^{N_m\times d}.
\end{equation}
In our implementation, $g_{\phi}$ uses strided 3D convolutions to downsample $\mathbf{s}$ to the same $(T',H',W')$ grid as the VAE latents, producing a token sequence by flattening the spatiotemporal grid (so $N_m=T'H'W'$).
We set the token channel dimension to match the DiT hidden size, allowing $\mathbf{m}$ to be injected as a first-class conditioning stream.

% ------------------------------------------------------------------------
\subsubsection{Conditioning interface in the DiT denoiser}
We condition the DiT denoiser on (i) motion tokens $\mathbf{m}$ and (ii) DINO appearance tokens $\mathbf{a}$.
Motion tokens provide spatiotemporally aligned guidance, while the DINO cross-attention branch injects localized appearance cues.
Overall, the denoiser operates as
\begin{equation}
\label{eq:full_cond}
\hat{\boldsymbol{\epsilon}}=
\epsilon_{\theta}\!\left(\mathbf{x}_\tau, \tau \mid \underbrace{\mathbf{m}}_{\text{motion}},\;
\underbrace{\mathbf{a}}_{\text{DINO-ALF}} \right),
\end{equation}
where $I_{\mathrm{ref}}$ influences generation only through the extracted tokens $\mathbf{a}$, and the pose sequence $\hat{\mathbf{P}}$ only through the rendered pose control $\mathbf{s}$ and its encoding $\mathbf{m}$.

% ------------------------------------------------------------------------

\subsubsection{Training and adaptation with LoRA}
Fully fine-tuning a large video DiT is expensive and prone to overfitting on limited complex-motion data.
We therefore keep the pretrained backbone frozen and train only the adaptation modules:
(i) LoRA modules in selected self-attention and MLP projections, (ii) the DINO cross-attention branch, (iii) the motion encoder $g_{\phi}$, and (iv) the DINO aggregation/projection heads ($A_{\gamma}$ and $p_{\eta}$).
For a linear projection $\mathbf{W}$ inside the DiT (e.g., $q,k,v,o$ or MLP layers), LoRA replaces
\begin{equation}
\label{eq:lora}
\mathbf{W}\mathbf{x}\;\;\mapsto\;\;(\mathbf{W} + \tfrac{\alpha}{r}\mathbf{B}\mathbf{A})\mathbf{x},
\end{equation}
where $\mathbf{A}\in\mathbb{R}^{r\times d_{\mathrm{in}}}$ and $\mathbf{B}\in\mathbb{R}^{d_{\mathrm{out}}\times r}$ are trainable low-rank matrices, $r$ is the rank, and $\alpha$ is a scaling factor.

% ------------------------------------------------------------------------
\subsubsection{Condition dropout for robustness}
To reduce over-reliance on any single conditioning channel and improve generalization, we apply dropout during training:
with small probabilities, we zero out the motion tokens $\mathbf{m}$ and/or the DINO appearance tokens $\mathbf{a}$.
This encourages the model to distribute responsibility between explicit motion control and appearance preservation, yielding more stable generation under challenging motion.

\paragraph{Inference.}
At test time, we (i) generate a skeleton sequence from text (Section~\ref{sec:text2pose}), (ii) rasterize it to a pose-control video $\mathbf{s}$ and encode it into motion tokens $\mathbf{m}$, (iii) extract DINO tokens $\mathbf{a}$ from $I_{\mathrm{ref}}$, and (iv) run the diffusion sampler in latent space with the trained conditioning modules.
Following~\cite{wananimate}, we apply a pose retargeting post-processing step that adjusts bone lengths and refines small joint positions to match the reference image before video synthesis. The final video is obtained by decoding the denoised latents using the pretrained VAE decoder.
% ------------------------------------------------------------------------

%\IEEEpubidadjcol

\section{Experiments}
\label{sec:experiments}

We evaluated our two-stage framework by first assessing text-to-skeleton generation quality, then pose-conditioned video synthesis, and finally reported ablation studies for key design choices in both stages.

\subsection{Datasets}
\label{sec:datasets}

\subsubsection{Text-Pose Dataset}
For text-to-skeleton evaluation, we used a total of 8{,}000 paired text--pose sequences drawn from:
(i) our in-house Blender-rendered complex-motion videos, and
(ii) the \emph{Fitness} category adopted by HumanDreamer~\cite{humandreamer} (sourced from Motion-X~\cite{motionx}).
Of these, 2{,}000 sequences originate from our Blender-rendered synthetic videos, with text prompts derived directly from the Mixamo motion names (e.g., a ``backflip'' motion is captioned as ``a person performs a backflip''), and the remainder from the Motion-X Fitness subset. We randomly split the data at the sequence level into 90\% for training and 10\% as a held-out test set, and reported all evaluation results on the test set. Each motion was represented as a view-aligned 2D pose sequence with $J{=}62$ joints before discretization as described in Section~\ref{sec:text2pose}.

\subsubsection{Synthetic Video Dataset}
To the best of our knowledge, there is no publicly available video dataset specifically focused on complex human movements suitable for video generation research. The TikTok dataset~\cite{tiktok_dance} is a publicly available copyright-free human video benchmark, but focuses on dance and upper-body movements rather than acrobatic actions. Human action recognition datasets such as UCF101~\cite{ucf_dataset}, HMDB51~\cite{hmdb_dataset}, KTH~\cite{kth_dataset}, and NTU RGB+D~\cite{ntu_dataset} are unsuitable for two reasons: they suffer from low resolution, compression artifacts, and visual noise that degrade fine-tuning of modern video generation models, and their action categories are limited to simple, isolated motions (e.g., clapping, jumping, waving) rather than complex acrobatic sequences such as backflips, cartwheels, and martial arts kicks. Prior datasets often rely on web-sourced videos~\cite{hypermotion, tiktok_dance}. Moreover, complex actions remain difficult even for advanced closed-source generators such as \textit{Sora}~\cite{sora} and \textit{Kling}~\cite{kling}, which makes them unreliable for building a high-quality benchmark in this regime. To avoid potential copyright and privacy concerns inherent to web-sourced data and to address these limitations, we constructed a synthetic dataset using \textit{Blender} with full control over characters, camera setting, and environments.

We curated a diverse set of complex-action FBX motions from Mixamo~\cite{mixamo} and pair them with Mixamo human characters and skins to increase appearance diversity. For scene variation, we used HDRI environment maps from PolyHaven~\cite{polyhaven} as backgrounds and lighting. We assembled each scene in Blender by placing the animated character, selecting camera setting, and rendering with consistent parameters. The resulting dataset contains 2{,}000 synthetic videos covering acrobatics and stunt-like motions across diverse characters and environments. This scale is larger than the official TikTok dance dataset (340 videos) and, being fully synthetic, does not introduce the privacy and consent issues of web-collected videos~\cite{tiktok_dance}.
Several representative samples of our constructed videos are provided on our \href{https://ashkantaghipour.github.io/kangaroo/}{project page} (Dataset section).

\begin{table*}[t]
  \centering
  \caption{Quantitative evaluation for Text-to-Skeleton generation on our 8{,}000-pair benchmark.
  $\downarrow$ indicates lower is better and $\uparrow$ indicates higher is better.}
  \vspace{-6pt}
  \label{tab:text2pose_metrics}
  \setlength{\tabcolsep}{4pt}
  \renewcommand{\arraystretch}{1.15}
  \begin{tabular}{lccccccc}
     \toprule
    Method & FID$\downarrow$ & Rp-top1$\uparrow$ & Rp-top2$\uparrow$ & Rp-top3$\uparrow$ & Diversity$\uparrow$ & MM-Dist$\downarrow$  \\
    \hline
    T2M-GPT~\cite{t2m_gpt}  & 524.61 & 0.191 & 0.287 & 0.473 & 40.11 & 49.85  \\
    PriorMDM~\cite{priormdm} & 585.31 & 0.216 & 0.325 & 0.501 & 42.58 & 44.29  \\
    MLD~\cite{mld}     & 467.22 & 0.335 & 0.503 & 0.653 & 41.67 & 47.66  \\
    HumanDreamer~\cite{humandreamer}      & 322.16 & 0.411 & 0.598 & 0.722 & 45.33 & 41.53  \\
    \hline
    \textbf{Ours} & \textbf{255.19} & \textbf{0.487} & \textbf{0.667} & \textbf{0.784} & \textbf{48.33} & \textbf{38.65}  \\
    \bottomrule
  \end{tabular}
\end{table*}

\subsection{Text-to-Skeleton Evaluation}
\label{sec:exp_text2pose}

\noindent \textbf{Baselines.}
We compared our method against HumanDreamer~\cite{humandreamer} using its released implementation, and we adopted the same baseline set and adaptation/evaluation protocol to ensure a consistent comparison with prior text-to-motion literature.
Specifically, we evaluated baselines including
T2M-GPT~\cite{t2m_gpt} (a two-stage VQ-VAE tokenizer followed by a GPT-style autoregressive transformer conditioned on text),
PriorMDM~\cite{priormdm} (a text-conditioned diffusion model that leverages a pretrained Motion Diffusion Model as a generative prior),
and MLD~\cite{mld} (a motion VAE paired with conditional latent diffusion in the learned motion-latent space).

\noindent \textbf{Evaluation metrics.} Following standard text-to-motion protocols~\cite{motion_metric_1, motion_metric_2, motion_metric_3, motion_metric_4}, all metrics were evaluated in a learned text--pose embedding space as defined in~\cite{humandreamer}. We reported:
(i) \textbf{FID} ($\downarrow$): measures the distributional similarity between generated and real motions to assess visual realism;
(ii) \textbf{R-precision} at top-$k$ ($k{=}{1,2,3}$; $\uparrow$): quantifies semantic accuracy by measuring how often the ground-truth text is correctly retrieved from a pool of distractors given a generated motion;
(iii) \textbf{Diversity} ($\uparrow$): measures the average geometric distance between generated samples to ensure a wide variety of actions across the dataset; and
(iv) \textbf{MM-Dist} ($\downarrow$): calculates the multimodal distance between text and motion features to reflect prompt adherence.

\noindent \textbf{Implementation details.}
We used tokenization with $K{=}256$ bins and offset $o{=}4$, and decoded with top-$k$ sampling ($k{=}10$). We trained using AdamW with learning rate $10^{-5}$, weight decay $0.01$, $\beta=(0.9, 0.95)$, and batch size 8 on a single NVIDIA A100 GPU.

% ------------------------------------------------------------------------

\begin{table*}[t]
    \centering
    \caption{Comparison of different methods on VBench.
    Higher is better for ($\uparrow$); lower is better for ($\downarrow$).}
    \vspace{-6pt}
    \label{tab:vbench_main}
    \scriptsize
    \setlength{\tabcolsep}{3.0pt}
    \renewcommand{\arraystretch}{1.05}
    \begin{tabular}{lccccccccccc}
        \toprule
        Model &
        \multicolumn{7}{c}{VBench-I2V} &
        \multicolumn{3}{c}{Frame-level} &
        \multicolumn{1}{c}{Video-level} \\
        \cmidrule(lr){2-8} \cmidrule(lr){9-11} \cmidrule(lr){12-12}
        &
        Subj. Cons$\uparrow$ &
        Bg. Cons$\uparrow$ &
        Motion Flick.$\uparrow$ &
        Motion Smooth.$\uparrow$ &
        Dyn.Degree$\uparrow$ &
        Aes.Qual$\uparrow$ &
        Img.Qual$\uparrow$ &
        SSIM$\uparrow$ &
        LPIPS$\downarrow$ &
        PSNR$\uparrow$ &
        FVD$\downarrow$ \\
        \midrule
        HumanVid~\cite{humanvid}                & 85.33 & 86.91 & 91.01 & 88.21 & 72.72 & 54.61 & 64.19 & 0.737 & 0.213 & 25.54 & 637.1 \\
        MimicMotion~\cite{mimicmotion}          & 86.01 & 88.47 & 92.11 & 89.95 & 73.05 & 55.07 & 65.14 & 0.744 & 0.198 & 26.66 & 611.5 \\
        Hyper-Motion~\cite{hypermotion}         & 86.66 & 90.23 & 93.22 & 90.08 & 73.65 & 55.21 & 65.42 & 0.767 & 0.181 & 27.13 & 568.8 \\
        UniAnimate-DiT~\cite{unianimate}        & 87.21 & 91.17 & 95.78 & 92.33 & 74.87 & 56.32 & 66.87 & 0.751 & 0.189 & 27.48 & 527.4 \\
        VACE~\cite{vace}                        & 88.31 & 91.11 & 95.31 & 94.81 & 75.03 & \textbf{57.11} & \textbf{67.91} & 0.771 & \textbf{0.173} & 27.91 & 498.7 \\
        \textbf{Ours}                           & \textbf{91.31} & \textbf{93.79} & \textbf{97.50} & \textbf{97.39} & \textbf{78.50} & 56.37 & 67.14 & \textbf{0.795} & 0.174 & \textbf{28.50} & \textbf{471.3} \\
        \bottomrule
    \end{tabular}
\end{table*}

\begin{table}[t]
  \centering
  \caption{Ablation study on the text-to-skeleton architecture.
  We analyze the effect of tokenization granularity ($K$), decoder depth ($L$), and decoding strategy.
  $\downarrow$ indicates lower is better; $\uparrow$ indicates higher is better.}
  \vspace{-6pt}
  \label{tab:ablation_text2pose}
  \setlength{\tabcolsep}{3pt}
  \renewcommand{\arraystretch}{1.15}
  \begin{tabular}{lccccc}
    \toprule
    Configuration & FID$\downarrow$ & Rp-top1$\uparrow$ & Diversity$\uparrow$ & MM-Dist$\downarrow$ \\
    \midrule
    \multicolumn{5}{l}{\textit{Tokenization granularity ($K$ bins)}} \\
    \quad $K=64$   & 312.41 & 0.401 & 44.21 & 43.17 \\
    \quad $K=128$  & 278.55 & 0.452 & 46.88 & 40.52 \\
    \quad $K=256$ (Baseline) & \textbf{255.19} & \textbf{0.487} & \textbf{48.33} & \textbf{38.65} \\
    \quad $K=512$  & 261.33 & 0.479 & 47.65 & 39.21 \\
    \midrule
    \multicolumn{5}{l}{\textit{Decoder depth ($L$ layers)}} \\
    \quad $L=6$   & 301.55 & 0.418 & 44.77 & 43.52 \\
    \quad $L=12$  & 272.38 & 0.459 & 46.91 & 40.33 \\
    \quad $L=18$ (Baseline) & \textbf{255.19} & \textbf{0.487} & \textbf{48.33} & \textbf{38.65} \\
    \quad $L=24$  & 258.91 & 0.481 & 47.88 & 39.01 \\
    \midrule
    \multicolumn{5}{l}{\textit{Decoding strategy}} \\
    \quad Greedy  & 289.66 & 0.441 & 42.15 & 41.77 \\
    \quad Nucleus ($p{=}0.9$) & 267.81 & 0.468 & 47.22 & 39.88 \\
    \quad Top-$k$ ($k{=}10$, Baseline) & \textbf{255.19} & \textbf{0.487} & \textbf{48.33} & \textbf{38.65} \\
    \bottomrule
  \end{tabular}
\end{table}

\noindent \textbf{Quantitative analysis.}
As shown in Table~\ref{tab:text2pose_metrics}, our approach outperformed existing methods across all evaluation metrics. Our model achieved an FID of 255.19, an improvement over the previous best of 322.16 held by HumanDreamer. This lower FID score indicated that our generated skeleton sequences more closely resembled the distribution of real human motions, suggesting improved physical plausibility and visual realism. Furthermore, our method showed the highest Diversity score (48.33), demonstrating that it could generate a wider range of distinct actions and avoided the common pitfall of mode collapse where a model produces repetitive movements.

In terms of semantic alignment and instruction following, the R-precision scores showed an advantage; our top-1 accuracy reached 0.487, meaning the correct text description is successfully retrieved as the best match nearly half the time within a pool of distractors. This is complemented by the Multimodal Distance (MM-Dist), where our model achieved the lowest score of 38.65. Together, these results indicated that our method developed a more precise mapping between the linguistic nuances of the input prompts and the resulting pose sequences. The consistent gains across Rp-top1, top-2, and top-3 further confirmed that our model's adherence to user prompts is both accurate and robust.

\begin{figure*}[t]
    \centering
    \includegraphics[width=\textwidth]{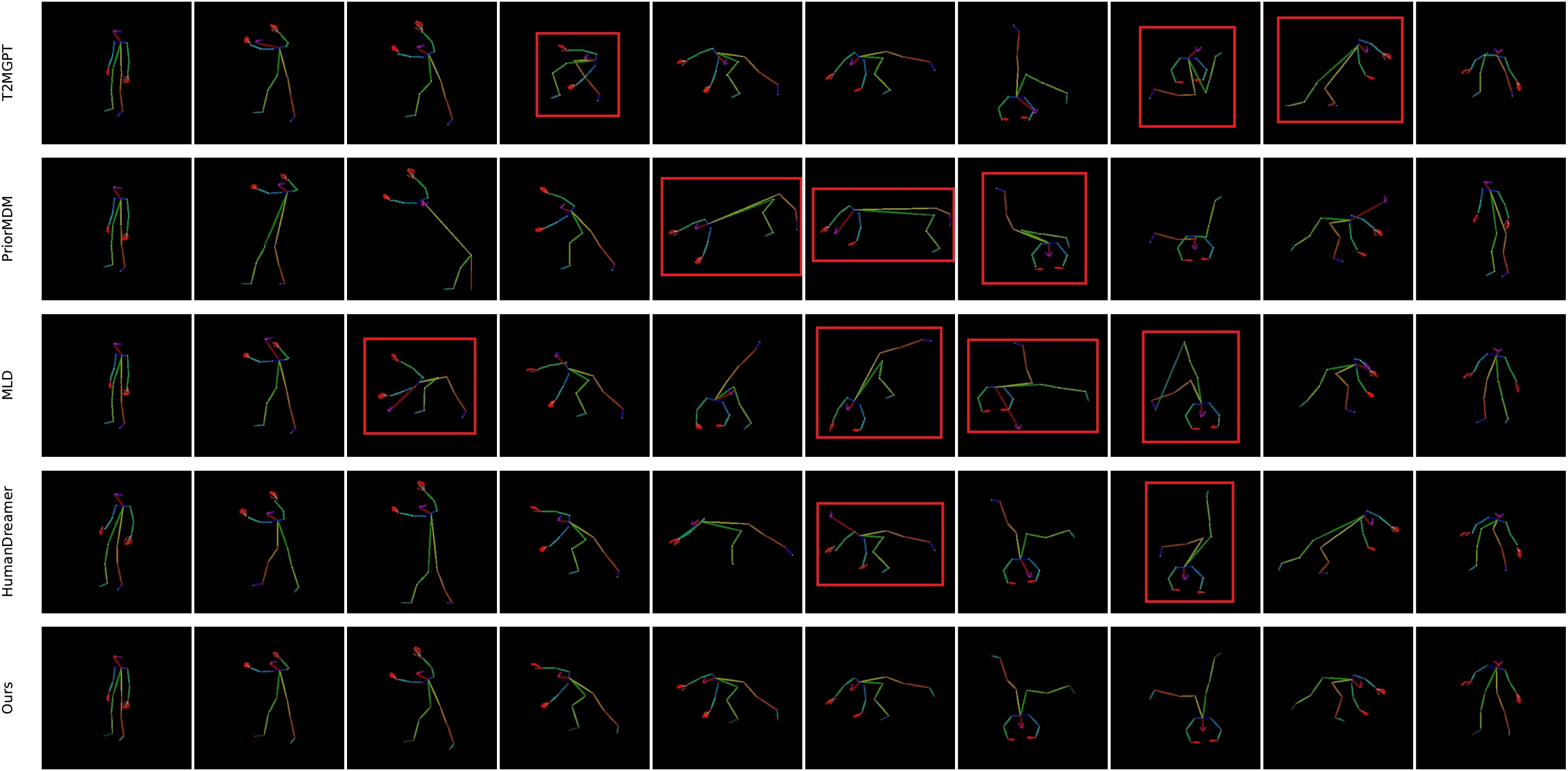}
\caption{Qualitative comparison of generated 2D pose sequences for the prompt ``A person performs a cartwheel''. Our method produces more coherent and physically plausible trajectories, especially during inverted phases. Failure cases of competing methods are highlighted with red boxes. For better visualization, we crop the scene to show only the human skeleton.}
    \vspace{-6pt}
    \label{fig:pose_methods}
\end{figure*}

\noindent \textbf{Qualitative analysis.}
Qualitative comparisons of generated pose sequences for the prompt ``A person performs a cartwheel'' are shown in Fig.~\ref{fig:pose_methods}. Our method produced more coherent and physically plausible trajectories, particularly during challenging inverted phases and rapid transitions. In contrast, HumanDreamer often produced implausible head and leg configurations, as highlighted by the red boxes. MLD also failed to generate plausible head and leg poses in several frames (red boxes). PriorMDM exhibited artifacts such as unrealistic leg poses and elongated body parts (red boxes), while T2M-GPT confused pose ordering and frequently produced distorted limbs (red boxes). Additional qualitative results for other prompts are provided on our \href{https://ashkantaghipour.github.io/kangaroo/}{project page}.

\subsection{Pose-to-Video Evaluation}
\label{sec:exp_pose2video}

\noindent\textbf{Baselines.}
We compared against SOTA pose-conditioned human video generation methods discussed in Section~\ref{sec:related}: HumanVid~\cite{humanvid}, MimicMotion~\cite{mimicmotion}, Hyper-Motion~\cite{hypermotion}, UniAnimate-DiT~\cite{unianimate}, and VACE~\cite{vace}. These methods were selected as they represent the main architectural paradigms for pose-conditioned generation, including ControlNet-style adapters, confidence-aware guidance, spatial RoPE design, and DiT-based conditioning.

\noindent\textbf{Implementation Details.}
We initialize our model from the pretrained Wan2.1~\cite{wan} I2V 14B video diffusion model and conduct all experiments on 4 NVIDIA A100 GPUs.
All videos contain 81 frames.
We fine-tune the model for 40{,}000 steps using a learning rate of $2\times 10^{-5}$.
Training is performed on our 2{,}000-video synthetic Blender dataset, and we adopt a parameter-efficient LoRA fine-tuning strategy with rank 64 (we fine-tune LoRA adapters and the additional conditioning modules introduced in Section~\ref{sec:pose2video}).
We randomly split the dataset at the video level into 1{,}800 videos for training and 200 for evaluation, ensuring no overlap between training and test data.

\noindent\textbf{Evaluation Metrics.}
We reported frame-level image quality metrics used in video generation, including SSIM~\cite{SSIM} (structural similarity), LPIPS~\cite{lpips} (perceptual distance in deep feature space), and PSNR~\cite{psnr} (pixel-wise reconstruction fidelity), as well as the video-level Fr\'echet Video Distance (FVD)~\cite{FVD}, which measures distributional distance between real and generated videos in a learned video feature space.
In addition, we used fine-grained VBench-I2V~\cite{vbench} metrics: \emph{Subject Consistency} (appearance preservation via pretrained visual feature similarity); \emph{Background Consistency} (background preservation via CLIP feature similarity); \emph{Temporal Flickering} (temporal flickering via frame differences); \emph{Motion Smoothness} (short-term dynamics via frame interpolation reconstruction error); \emph{Dynamic Degree} (proportion of non-static videos via RAFT~\cite{raft} optical-flow); \emph{Aesthetic Quality} (LAION aesthetic scores); and \emph{Imaging Quality} (MUSIQ~\cite{musiq} scores for low-level distortions).

\noindent \textbf{Quantitative Results.}
Table~\ref{tab:vbench_main} compared our method against SOTA baselines on VBench-I2V metrics, frame-level quality measures, and video-level FVD. Our method achieved the best performance on five of seven VBench-I2V metrics, with comparable scores on the remaining two. Specifically, we outperformed the strongest baseline (VACE) by +3.0 on \emph{Subject Consistency} (91.31 vs.\ 88.31), +2.68 on \emph{Background Consistency} (93.79 vs.\ 91.11), and +2.58 on \emph{Motion Smoothness} (97.39 vs.\ 94.81). These improvements indicated that our DINO-ALF appearance encoding better preserves subject identity and background coherence, while structured pose conditioning yields smoother motion trajectories. The +3.47 gain on \emph{Dynamic Degree} (78.50 vs.\ 75.03) further confirmed that our method produced videos with richer, non-static motion rather than trivial static solutions.

On frame-level metrics, we achieved the highest SSIM (0.795) and PSNR (28.50), with competitive LPIPS (0.174 vs.\ VACE's 0.173). At the video level, our FVD of 471.3 is the lowest, indicating that generated videos better match the real-video distribution. While VACE attains slightly higher \emph{Aesthetic Quality} (57.11 vs.\ 56.37) and \emph{Imaging Quality} (67.91 vs.\ 67.14), our method delivered substantially stronger temporal consistency and motion realism. These results demonstrated that our cascade design—decoupling motion control from appearance synthesis—improves motion dynamics without sacrificing visual quality.

\begin{figure*}[t]
    \centering
    \includegraphics[width=\textwidth]{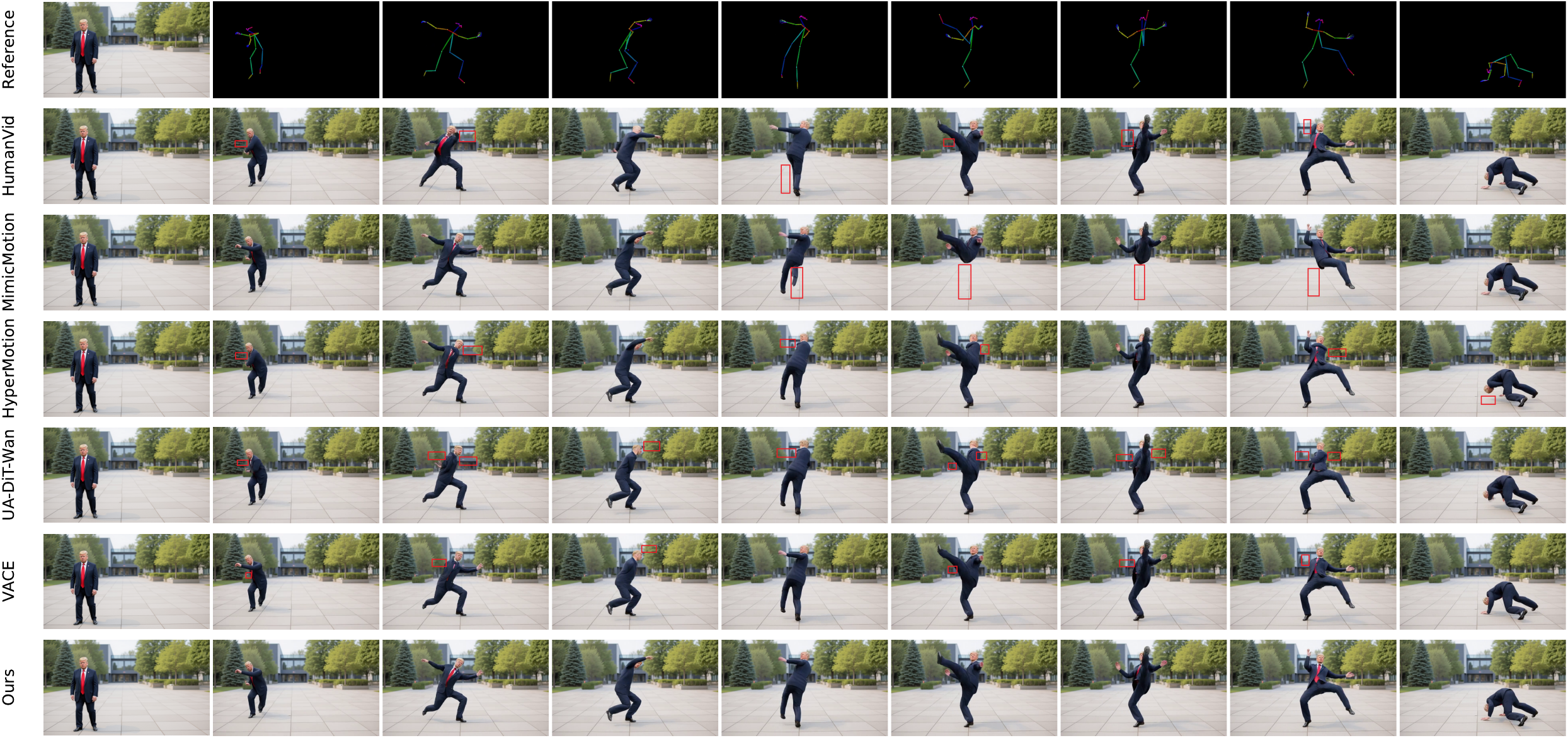}
\caption{Qualitative video comparison with pose-guided baselines. Our method best follows the target 2D skeleton while preserving the reference appearance. Competing methods exhibit hand/leg artifacts under complex motions (red boxes), and our DINO-ALF conditioning better retains fine details (e.g., the red tie). More results are on our \href{https://ashkantaghipour.github.io/kangaroo/}{project page}.}
    \vspace{-6pt}
    \label{fig:my_figure_compare}
\end{figure*}

% ============================================================
% ABLATION TABLE 2: Pose-to-Video Network Parameters
% ============================================================

\begin{table}[t]
    \centering
    \caption{Ablation study on pose-to-video network parameters.
    We analyze the effect of LoRA rank, motion encoder architecture, and condition dropout rate.
    Higher is better for ($\uparrow$); lower is better for ($\downarrow$).}
    \vspace{-6pt}
    \label{tab:ablation_pose2video}
    \setlength{\tabcolsep}{1.5pt}
    \renewcommand{\arraystretch}{1.10}
    \scriptsize
    \begin{tabular}{lcccccc}
        \toprule
        Configuration & Subj.$\uparrow$ & Bg.$\uparrow$ & M.Smooth$\uparrow$ & SSIM$\uparrow$ & LPIPS$\downarrow$ & FVD$\downarrow$ \\
        \midrule
        \multicolumn{7}{l}{\textit{LoRA rank ($r$)}} \\
        \quad $r=16$  & 84.22 & 90.45 & 92.11 & 0.712 & 0.221 & 512.8 \\
        \quad $r=32$  & 88.77 & 91.58 & 95.88 & 0.758 & 0.193 & 491.5 \\
        \quad $r=64$ (Baseline) & 91.31 & \textbf{93.79} & \textbf{97.39} & \textbf{0.795} & \textbf{0.174} & \textbf{471.3} \\
        \quad $r=128$ & \textbf{91.33} & 93.72 & 97.28 & 0.789 & 0.178 & 478.9 \\
        \midrule
        \multicolumn{7}{l}{\textit{Motion encoder architecture}} \\
        \quad 2D CNN + temporal attn. & 87.91 & 90.55 & 93.22 & 0.755 & 0.195 & 521.4 \\
        \quad 3D CNN shallow (4 layers) & 89.22 & 91.88 & 95.11 & 0.771 & 0.185 & 498.2 \\
        \quad 3D CNN deep (8 layers, Baseline) & \textbf{91.31} & \textbf{93.79} & \textbf{97.39} & \textbf{0.795} & \textbf{0.174} & \textbf{471.3} \\
        \midrule
        \multicolumn{7}{l}{\textit{Condition dropout rate ($p_{\text{drop}}$)}} \\
        \quad $p_{\text{drop}}=0$ (no dropout) & 89.11 & 92.05 & 95.22 & 0.775 & 0.182 & 492.1 \\
        \quad $p_{\text{drop}}=0.05$ & 90.22 & 92.88 & 96.33 & 0.785 & 0.178 & 481.7 \\
        \quad $p_{\text{drop}}=0.1$ (Baseline) & \textbf{91.31} & \textbf{93.79} & \textbf{97.39} & \textbf{0.795} & \textbf{0.174} & \textbf{471.3} \\
        \quad $p_{\text{drop}}=0.2$ & 88.55 & 91.13 & 95.81 & 0.769 & 0.191 & 501.5 \\
        \bottomrule
    \end{tabular}
\end{table}

\noindent \textbf{Qualitative Results.}
Fig.~\ref{fig:my_figure_compare} presents visual comparisons between our method and the baselines. Our method closely followed the target 2D skeleton while preserving the reference appearance across diverse motions. In contrast, all baselines exhibited artifacts under complex poses. Hand generation is particularly challenging: HumanVid~\cite{humanvid} produced incorrect hands in early and late frames, Hyper-Motion~\cite{hypermotion} failed during spinning movements, UniAnimate-DiT~\cite{unianimate} missed parts of both hands, and VACE~\cite{vace} incorrectly rendered the right hand (all highlighted by red boxes). MimicMotion~\cite{mimicmotion} additionally missed one leg during the kicking motion. Our method avoided these failures by leveraging DINO-ALF, which better attends to fine-grained appearance details—for example, the red tie is consistently preserved (compare third and eighth columns). Additional results are available on our \href{https://ashkantaghipour.github.io/kangaroo/}{project page}.

\noindent\textbf{Ablation Study.}
We ablated key architectural and inference choices in Table~\ref{tab:ablation_text2pose}, examining tokenization granularity ($K$), decoder depth ($L$), and decoding strategy.

\noindent\textbf{(1) Tokenization granularity.}
Increasing $K$ improves realism and text-motion alignment by reducing discretization error, but overly fine tokenization slightly degrades performance due to a larger, sparser vocabulary that is harder to model. We use $K{=}256$.

\noindent\textbf{(2) Decoder depth.}
Deeper decoders better capture long-range temporal structure and global body coordination, with gains saturating at larger depths. We use $L{=}18$.

\noindent\textbf{(3) Decoding strategy.}
Greedy decoding (argmax per step) is overly deterministic and reduces diversity, often propagating early errors. Stochastic sampling (nucleus / Top-$k$) yields a better quality--diversity trade-off; we use Top-$k$ with $k{=}10$.

We also conducted ablation studies on the pose-to-video generation stage to analyze key design choices. We first ablated the appearance encoding design in Table~\ref{tab:ablation}, then analyzed network parameters in Table~\ref{tab:ablation_pose2video}.

\begin{table*}[t]
    \centering
    \caption{Ablation study on appearance encoding design. We compare DINO-ALF (Ours), DLL (single-layer), CLIP embeddings, and no appearance encoder.
    Higher is better for ($\uparrow$); lower is better for ($\downarrow$).}
    \vspace{-6pt}
    \label{tab:ablation}
    \scriptsize
    \setlength{\tabcolsep}{3.0pt}
    \renewcommand{\arraystretch}{1.05}
    \begin{tabular}{lccccccccccc}
        \toprule
        Model &
        \multicolumn{7}{c}{VBench-I2V} &
        \multicolumn{3}{c}{Frame-level} &
        \multicolumn{1}{c}{Video-level} \\
        \cmidrule(lr){2-8} \cmidrule(lr){9-11} \cmidrule(lr){12-12}
        &
        Subj. Cons$\uparrow$ &
        Bg. Cons$\uparrow$ &
        Motion Flick.$\uparrow$ &
        Motion Smooth.$\uparrow$ &
        Dyn.Degree$\uparrow$ &
        Aes.Qual$\uparrow$ &
        Img.Qual$\uparrow$ &
        SSIM$\uparrow$ &
        LPIPS$\downarrow$ &
        PSNR$\uparrow$ &
        FVD$\downarrow$ \\
        \midrule
        DLL (single-layer)      & 87.65 & 90.03 & 94.11 & 93.77 & 74.13 & 55.11 & 66.23 & 0.752 & 0.181 & 27.02 & 502.5 \\
        CLIP embeddings         & 85.44 & 89.75 & 93.63 & 91.21 & 73.02 & 54.76 & 66.08 & 0.741 & 0.195 & 26.98 & 517.7 \\
        No appearance encoder   & 75.13 & 85.14 & 82.22 & 82.11 & 65.53 & 47.11 & 60.09 & 0.673 & 0.245 & 24.33 & 593.4 \\
        \textbf{Ours}   & \textbf{91.31} & \textbf{93.79} & \textbf{97.50} & \textbf{97.39} & \textbf{78.50} & \textbf{56.37} & \textbf{67.14} & \textbf{0.795} & \textbf{0.174} & \textbf{28.50} & \textbf{471.3} \\
        \bottomrule
    \end{tabular}
\end{table*}

\begin{figure*}[t]
    \centering
    \includegraphics[width=\textwidth]{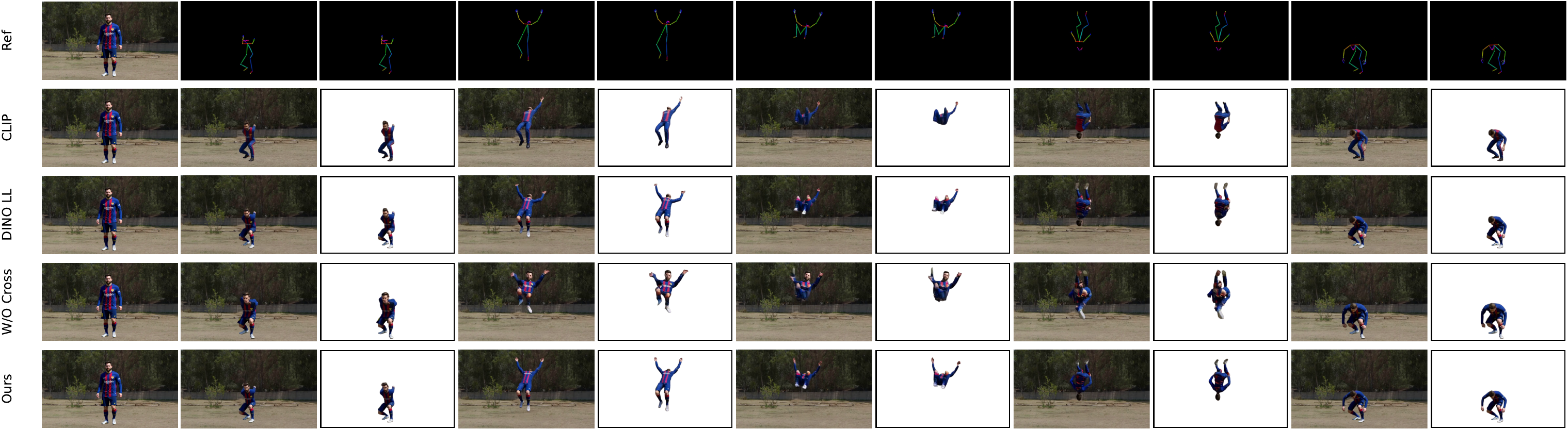}
\caption{Ablation study on the pose-to-video stage. Rows from top to bottom: Ref (reference frame and target pose), CLIP (CLIP embeddings), DINO LL (DLL single-layer), W/O cross (no appearance encoder), and Ours. DINO-ALF yields the best appearance preservation under large pose changes, while CLIP-only or no appearance encoder causes hand/clothing artifacts and identity drift (red boxes).}
    \vspace{-6pt}
    \label{fig:ablation_video}
\end{figure*}

\noindent\textbf{Effect of DINO-ALF.}
We first studied the impact of using DINO-ALF for reference appearance conditioning. Instead of aggregating features from multiple DINO layers, we evaluated a common variant that uses only the DINOv3 last-layer (DLL) features. Compared to this variant, DINO-ALF consistently improved subject preservation under large pose changes. As shown in Fig.~\ref{fig:ablation_video} (third row, DINO LL), fine motion details—especially small hands (red boxes)—were missed when only DLL was used. This confirmed that combining low-level texture cues with high-level semantic features was crucial for robust appearance consistency. The quantitative results in Table~\ref{tab:ablation} further supported this observation: DINO-ALF improved subject consistency and motion-related metrics (motion smoothness and motion flickering), indicating more stable reference injection.

\noindent\textbf{Replacing DINO features with CLIP image embeddings.}
Next, we replaced the proposed DINO-based appearance encoder with the default CLIP image embeddings used in the original Wan2.1 architecture, injected via standard cross-attention. This variant showed noticeable degradation in fine-grained identity details under complex motions (Fig.~\ref{fig:ablation_video}, second row CLIP, red boxes), including missing hand parts and clothing drift (e.g., shorts turning into pants) as well as changes in shoe color (white to black). These results suggested that CLIP features alone were insufficient for preserving appearance under fast, complex motions. This was also reflected in Table~\ref{tab:ablation}, where CLIP underperformed DINO-ALF on subject consistency and motion metrics (motion smoothness, dynamic degree, and motion flickering), as well as frame-level metrics (SSIM, LPIPS).

\noindent\textbf{Removing the appearance encoder.}
We ablated the appearance encoder entirely by disabling the additional cross-attention branch, leaving only the pretrained reference pathway. This setting led to severe identity drift and unstable appearance across frames. As highlighted in Fig.~\ref{fig:ablation_video} (fourth row W/O cross, red boxes), during the backflip the generated video no longer followed the reference 2D pose reliably. Table~\ref{tab:ablation} also showed clear drops in aesthetics, motion, and subject consistency metrics, demonstrating that DINO-ALF was essential for reliable pose-driven video generation.

\noindent\textbf{Network parameter sensitivity.}
Table~\ref{tab:ablation_pose2video} ablated LoRA rank, motion encoder architecture, and condition dropout. For LoRA rank, performance improved from $r{=}16$ to $r{=}64$ (Subject Consistency: 84.22 → 91.31), but saturated at $r{=}128$. For motion encoding, the deep 3D CNN outperformed 2D CNN with temporal attention (FVD: 471.3 vs.\ 521.4), as it captured local spatiotemporal dynamics aligned with the latent grid. For condition dropout, $p_{\text{drop}}{=}0.1$ achieved the best balance; no dropout caused over-reliance on one conditioning stream, while $p_{\text{drop}}{=}0.2$ under-conditioned training. We used $r{=}64$, deep 3D CNN, and $p_{\text{drop}}{=}0.1$ in all experiments.

\noindent\textbf{Pose augmentation for error robustness.}
Table~\ref{tab:ablation_poseaug} evaluated the effect of pose augmentation when the video model was driven by predicted (rather than ground-truth) skeletons. Without augmentation, the video model was sensitive to artifacts in the generated skeletons, as it was trained exclusively on clean ground-truth poses. Each augmentation targeted a specific failure mode of the skeleton generation stage: joint jitter mitigated spatial coordinate noise, joint dropout handled missed or undetected joints, and temporal shift addressed frame-level timing errors. Combining all three yielded the best results, confirming that training-time augmentation was essential for robust cascaded generation. To isolate errors introduced by the skeleton generation stage, Table~\ref{tab:ablation_poseaug} also includes a GT skeleton upper bound (taken from Table~\ref{tab:vbench_main}); the gap between GT and full augmentation quantifies the cost of using predicted skeletons.

\begin{table}[t]
    \centering
    \caption{Ablation study on pose augmentation for error robustness. All rows except the first use \emph{predicted} skeletons at inference; the GT row serves as an upper bound.
    Higher is better for ($\uparrow$); lower is better for ($\downarrow$).}
    \vspace{-6pt}
    \label{tab:ablation_poseaug}
    \setlength{\tabcolsep}{2.5pt}
    \renewcommand{\arraystretch}{1.10}
    \scriptsize
    \begin{tabular}{lcccccc}
        \toprule
        Configuration & Subj.$\uparrow$ & Bg.$\uparrow$ & M.Smooth$\uparrow$ & SSIM$\uparrow$ & LPIPS$\downarrow$ & FVD$\downarrow$ \\
        \midrule
        GT skeletons (upper bound) & 91.31 & 93.79 & 97.39 & 0.795 & 0.174 & 471.3 \\
        \midrule
        No augmentation            & 85.11 & 86.57 & 89.07 & 0.652 & 0.229 & 574.7 \\
        Joint jitter only          & 86.45 & 87.26 & 91.39 & 0.651 & 0.213 & 568.1 \\
        Joint dropout only         & 86.57 & 87.73 & 91.11 & 0.683 & 0.208 & 541.5 \\
        Temporal shift only        & 87.31 & 88.05 & 92.71 & 0.713 & 0.193 & 528.3 \\
        Full augmentation (Ours)   & 89.23 & 91.17 & 95.61 & 0.781 & 0.183 & 492.5 \\
        \bottomrule
    \end{tabular}
\end{table}

\section{Limitations}
The current framework is designed for single-person motion generation and does not model multi-person interactions, such as coordinated group actions, physical contact, or collision avoidance. Extending to multi-person scenarios would require jointly modeling inter-person spatial relationships and their interactions, which we leave for future work. Additionally, under very fast rotations and acrobatic transitions, fine-grained details such as fingers and facial features may be lost or blurred, as the pose-conditioned model struggles to preserve high-frequency appearance cues at these extremities.

\section{Conclusion}
We presented a cascaded framework for controllable complex human motion video generation that decouples motion planning from appearance synthesis. An autoregressive text-to-skeleton model generates 2D pose sequences from natural language, while a pose-conditioned video diffusion model with DINO-ALF preserves appearance under large deformations and self-occlusions. We also introduced a synthetic benchmark of 2,000 complex-motion videos addressing the under-representation of acrobatic actions in existing datasets. Experiments demonstrated that our framework outperformed prior methods on both text-to-skeleton and video generation metrics.

% or


% \section*{Acknowledgment}


\ifCLASSOPTIONcaptionsoff
  \newpage
\fi

\bibliographystyle{IEEEtran}   % or the style required by TIP
\bibliography{bare_jrnl}       % without the .bib extension

@String(CVPR= {IEEE Conf. Comput. Vis. Pattern Recog.})

@String(ECCV= {Eur. Conf. Comput. Vis.})

@String(ICPR = {Int. Conf. Pattern Recog.})

@String(ICLR = {Int. Conf. Learn. Represent.})

@String(AAAI = {AAAI})

@String(CVPR  = {CVPR})

@String(ECCV  = {ECCV})

@String(ICPR  = {ICPR})

@String(ICLR  = {ICLR})

@inproceedings{humandreamer,
  title={HumanDreamer: Generating Controllable Human-Motion Videos via Decoupled Generation},
  author={Wang, Boyuan and Wang, Xiaofeng and Ni, Chaojun and Zhao, Guosheng and Yang, Zhiqin and Zhu, Zheng and Zhang, Muyang and Zhou, Yukun and Chen, Xinze and Huang, Guan and others},
  booktitle={Proceedings of the Computer Vision and Pattern Recognition Conference},
  pages={12391--12401},
  year={2025}
}

@article{2d_skelet,
  title={Toward Rich Video Human-Motion2D Generation},
  author={Xi, Ruihao and Wang, Xuekuan and Li, Yongcheng and Li, Shuhua and Wang, Zichen and Wang, Yiwei and Wei, Feng and Zhao, Cairong},
  journal={arXiv preprint arXiv:2506.14428},
  year={2025}
}

@article{wan,
  title={Wan: Open and advanced large-scale video generative models},
  author={Wan, Team and Wang, Ang and Ai, Baole and Wen, Bin and Mao, Chaojie and Xie, Chen-Wei and Chen, Di and Yu, Feiwu and Zhao, Haiming and Yang, Jianxiao and others},
  journal={arXiv preprint arXiv:2503.20314},
  year={2025}
}

@article{cogvideox,
  title={Cogvideox: Text-to-video diffusion models with an expert transformer},
  author={Yang, Zhuoyi and Teng, Jiayan and Zheng, Wendi and Ding, Ming and Huang, Shiyu and Xu, Jiazheng and Yang, Yuanming and Hong, Wenyi and Zhang, Xiaohan and Feng, Guanyu and others},
  journal={arXiv preprint arXiv:2408.06072},
  year={2024}
}

@article{hunyuanvideo,
  title={HunyuanVideo 1.5 Technical Report},
  author={Wu, Bing and Zou, Chang and Li, Changlin and Huang, Duojun and Yang, Fang and Tan, Hao and Peng, Jack and Wu, Jianbing and Xiong, Jiangfeng and Jiang, Jie and others},
  journal={arXiv preprint arXiv:2511.18870},
  year={2025}
}

@misc{sora,
  author       = {OpenAI},
  title        = {Sora: A Large-Scale Text-to-Video Model},
  howpublished = {\url{https://sora.chatgpt.com}},
  note         = {Accessed: 2025-12-11}
}

@misc{kling,
  author       = {Kuaishou Technology},
  title        = {KLING: High-Fidelity Text-to-Video Generation System},
  howpublished = {\url{https://klingai.com/global/}},
  note         = {Accessed: 2025-12-11}
}

@article{vedaldi,
  title={What Happens Next? Anticipating Future Motion by Generating Point Trajectories},
  author={Boduljak, Gabrijel and Karazija, Laurynas and Laina, Iro and Rupprecht, Christian and Vedaldi, Andrea},
  journal={arXiv preprint arXiv:2509.21592},
  year={2025}
}

@article{steadydancer,
  title={SteadyDancer: Harmonized and Coherent Human Image Animation with First-Frame Preservation},
  author={Zhang, Jiaming and Cao, Shengming and Li, Rui and Zhao, Xiaotong and Cui, Yutao and Hou, Xinglin and Wu, Gangshan and Chen, Haolan and Xu, Yu and Wang, Limin and others},
  journal={arXiv preprint arXiv:2511.19320},
  year={2025}
}

@article{holistic_motion_2d,
  title={Holistic-motion2d: Scalable whole-body human motion generation in 2d space},
  author={Wang, Yuan and Wang, Zhao and Gong, Junhao and Huang, Di and He, Tong and Ouyang, Wanli and Jiao, Jile and Feng, Xuetao and Dou, Qi and Tang, Shixiang and others},
  journal={arXiv preprint arXiv:2406.11253},
  year={2024}
}

@article{motion_2d_to_3d,
  title={Motion-2-to-3: Leveraging 2d motion data to boost 3d motion generation},
  author={Pi, Huaijin and Guo, Ruoxi and Shen, Zehong and Shuai, Qing and Hu, Zechen and Wang, Zhumei and Dong, Yajiao and Hu, Ruizhen and Komura, Taku and Peng, Sida and others},
  journal={arXiv preprint arXiv:2412.13111},
  year={2024}
}

@article{scail,
  title={SCAIL: Towards Studio-Grade Character Animation via In-Context Learning of 3D-Consistent Pose Representations},
  author={Yan, Wenhao and Ye, Sheng and Yang, Zhuoyi and Teng, Jiayan and Dong, ZhenHui and Wen, Kairui and Gu, Xiaotao and Liu, Yong-Jin and Tang, Jie},
  journal={arXiv preprint arXiv:2512.05905},
  year={2025}
}

@article{wananimate,
  title={Wan-animate: Unified character animation and replacement with holistic replication},
  author={Cheng, Gang and Gao, Xin and Hu, Li and Hu, Siqi and Huang, Mingyang and Ji, Chaonan and Li, Ju and Meng, Dechao and Qi, Jinwei and Qiao, Penchong and others},
  journal={arXiv preprint arXiv:2509.14055},
  year={2025}
}

@inproceedings{signllm,
  title={Signllm: Sign language production large language models},
  author={Fang, Sen and Chen, Chen and Wang, Lei and Zheng, Ce and Sui, Chunyu and Tian, Yapeng},
  booktitle={Proceedings of the IEEE/CVF International Conference on Computer Vision},
  pages={6622--6634},
  year={2025}
}

@inproceedings{signidd,
  title={Sign-idd: Iconicity disentangled diffusion for sign language production},
  author={Tang, Shengeng and He, Jiayi and Guo, Dan and Wei, Yanyan and Li, Feng and Hong, Richang},
  booktitle={Proceedings of the AAAI Conference on Artificial Intelligence},
  volume={39},
  number={7},
  pages={7266--7274},
  year={2025}
}

@inproceedings{signmix,
  title={Mixed signals: Sign language production via a mixture of motion primitives},
  author={Saunders, Ben and Camgoz, Necati Cihan and Bowden, Richard},
  booktitle={Proceedings of the IEEE/CVF International Conference on Computer Vision},
  pages={1919--1929},
  year={2021}
}

@article{Mimic2DM,
  title={Learning to Control Physically-simulated 3D Characters via Generating and Mimicking 2D Motions},
  author={Li, Jianan and Chen, Xiao and Huang, Tao and Wong, Tien-Tsin},
  journal={arXiv preprint arXiv:2512.08500},
  year={2025}
}

@article{survey_sign,
  title={A comprehensive survey on human video generation: Challenges, methods, and insights},
  author={Lei, Wentao and Wang, Jinting and Ma, Fengji and Huang, Guanjie and Liu, Li},
  journal={arXiv preprint arXiv:2407.08428},
  year={2024}
}

@article{motionx,
  title={Motion-x++: A large-scale multimodal 3d whole-body human motion dataset},
  author={Zhang, Yuhong and Lin, Jing and Zeng, Ailing and Wu, Guanlin and Lu, Shunlin and Fu, Yurong and Cai, Yuanhao and Zhang, Ruimao and Wang, Haoqian and Zhang, Lei},
  journal={arXiv preprint arXiv:2501.05098},
  year={2025}
}

@article{gendop,
  title={GenDoP: Auto-regressive Camera Trajectory Generation as a Director of Photography},
  author={Zhang, Mengchen and Wu, Tong and Tan, Jing and Liu, Ziwei and Wetzstein, Gordon and Lin, Dahua},
  journal={arXiv preprint arXiv:2504.07083},
  year={2025}
}

@inproceedings{t2m_gpt,
  title={Generating human motion from textual descriptions with discrete representations},
  author={Zhang, Jianrong and Zhang, Yangsong and Cun, Xiaodong and Zhang, Yong and Zhao, Hongwei and Lu, Hongtao and Shen, Xi and Shan, Ying},
  booktitle={Proceedings of the IEEE/CVF conference on computer vision and pattern recognition},
  pages={14730--14740},
  year={2023}
}

@inproceedings{priormdm,
  title={Human Motion Diffusion as a Generative Prior},
  author={Shafir, Yoni and Tevet, Guy and Kapon, Roy and Bermano, Amit Haim},
  year={2024},
  booktitle={The Twelfth International Conference on Learning Representations}
}

@inproceedings{mld,
  title={Executing your Commands via Motion Diffusion in Latent Space},
  author={Chen, Xin and Jiang, Biao and Liu, Wen and Huang, Zilong and Fu, Bin and Chen, Tao and Yu, Gang},
  booktitle={Proceedings of the IEEE/CVF Conference on Computer Vision and Pattern Recognition},
  pages={18000--18010},
  year={2023}
}

@inproceedings{motion_metric_1,
  title={Executing your commands via motion diffusion in latent space},
  author={Chen, Xin and Jiang, Biao and Liu, Wen and Huang, Zilong and Fu, Bin and Chen, Tao and Yu, Gang},
  booktitle={Proceedings of the IEEE/CVF conference on computer vision and pattern recognition},
  pages={18000--18010},
  year={2023}
}

@inproceedings{motion_metric_2,
  title={Temos: Generating diverse human motions from textual descriptions},
  author={Petrovich, Mathis and Black, Michael J and Varol, G{\"u}l},
  booktitle={European Conference on Computer Vision},
  pages={480--497},
  year={2022},
  organization={Springer}
}

@ARTICLE{motion_metric_3,
  author={Wang, Xinghan and Kang, Zixi and Mu, Yadong},
  journal={IEEE Transactions on Image Processing}, 
  title={Text-Controlled Motion Mamba: Text-Instructed Temporal Grounding of Human Motion}, 
  year={2025},
  volume={34},
  number={},
  pages={7079-7092},
  doi={10.1109/TIP.2025.3624601}}

@article{motion_metric_4,
  title={Human motion diffusion model},
  author={Tevet, Guy and Raab, Sigal and Gordon, Brian and Shafir, Yonatan and Cohen-Or, Daniel and Bermano, Amit H},
  journal={arXiv preprint arXiv:2209.14916},
  year={2022}
}

@article{humanvid,
  title={Humanvid: Demystifying training data for camera-controllable human image animation},
  author={Wang, Zhenzhi and Li, Yixuan and Zeng, Yanhong and Fang, Youqing and Guo, Yuwei and Liu, Wenran and Tan, Jing and Chen, Kai and Xue, Tianfan and Dai, Bo and others},
  journal={Advances in Neural Information Processing Systems},
  volume={37},
  pages={20111--20131},
  year={2024}
}

@inproceedings{mimicmotion,
  title={MimicMotion: High-Quality Human Motion Video Generation with Confidence-aware Pose Guidance},
  author={Yuang Zhang and Jiaxi Gu and Li-Wen Wang and Han Wang and Junqi Cheng and Yuefeng Zhu and Fangyuan Zou},
  booktitle={International Conference on Machine Learning},
  year={2025}
}

@article{hypermotion,
  title={HyperMotion: DiT-Based Pose-Guided Human Image Animation of Complex Motions},
  author={Xu, Shuolin and Zheng, Siming and Wang, Ziyi and Yu, HC and Chen, Jinwei and Zhang, Huaqi and Li, Bo and Jiang, Peng-Tao},
  journal={arXiv preprint arXiv:2505.22977},
  year={2025}
}

@article{unianimate,
      title={UniAnimate: Taming Unified Video Diffusion Models for Consistent Human Image Animation},
      author={Wang, Xiang and Zhang, Shiwei and Gao, Changxin and Wang, Jiayu and Zhou, Xiaoqiang and Zhang, Yingya and Yan, Luxin and Sang, Nong},
      journal={Science China Information Sciences},
      year={2025}
}

@inproceedings{vace,
    title = {VACE: All-in-One Video Creation and Editing},
    author = {Jiang, Zeyinzi and Han, Zhen and Mao, Chaojie and Zhang, Jingfeng and Pan, Yulin and Liu, Yu},
    booktitle = {Proceedings of the IEEE/CVF International Conference on Computer Vision},
    pages = {17191-17202},
    year = {2025}
}

@article{xu2023ViTPose++,
  title={ViTPose++: Vision Transformer Foundation Model for Generic Body Pose Estimation},
  author={Xu, Yufei and Zhang, Jing and Zhang, Qiming and Tao, Dacheng},
  journal={IEEE Transactions on Pattern Analysis and Machine Intelligence},
  year={2024},
  volume={46},
  pages={1212-1230},
  doi={10.1109/TPAMI.2023.3330016}
}

@inproceedings{sdv2,
  title={High-resolution image synthesis with latent diffusion models},
  author={Rombach, Robin and Blattmann, Andreas and Lorenz, Dominik and Esser, Patrick and Ommer, Bj{\"o}rn},
  booktitle={Proceedings of the IEEE/CVF conference on computer vision and pattern recognition},
  pages={10684--10695},
  year={2022}
}

@misc{mixamo,
  author       = {{Mixamo}},
  title        = {Mixamo: 3D Characters and Animations},
  howpublished = {\url{https://www.mixamo.com/}},
  note         = {Accessed: 2026-01-03}
}

@misc{polyhaven,
  author       = {{Poly Haven}},
  title        = {Poly Haven: Public Domain 3D Assets (HDRIs, Textures, and Models)},
  howpublished = {\url{https://polyhaven.com/}},
  note         = {Accessed: 2026-01-03}
}

@InProceedings{tiktok_dance,
    author    = {Jafarian, Yasamin and Park, Hyun Soo},
    title     = {Learning High Fidelity Depths of Dressed Humans by Watching Social Media Dance Videos},
    booktitle = {Proceedings of the IEEE/CVF Conference on Computer Vision and Pattern Recognition (CVPR)},
    month     = {June},
    year      = {2021},
    pages     = {12753-12762}
}

@article{echomotion,
  title={EchoMotion: Unified Human Video and Motion Generation via Dual-Modality Diffusion Transformer},
  author={Yang, Yuxiao and Sheng, Hualian and Cai, Sijia and Lin, Jing and Wang, Jiahao and Deng, Bing and Lu, Junzhe and Wang, Haoqian and Ye, Jieping},
  journal={arXiv preprint arXiv:2512.18814},
  year={2025}
}

@article{4d_animation,
  title={MTVCraft: Tokenizing 4D Motion for Arbitrary Character Animation},
  author={ControlNeXt, MimicMotion and UniAnimate, StableAnimator and UniAnimate, StableAnimator}
}

@article{one_to_all,
  title={One-to-All Animation: Alignment-Free Character Animation and Image Pose Transfer},
  author={Shi, Shijun and Xu, Jing and Li, Zhihang and Peng, Chunli and Yang, Xiaoda and Lu, Lijing and Hu, Kai and Zhang, Jiangning},
  journal={arXiv preprint arXiv:2511.22940},
  year={2025}
}

@inproceedings{champ,
  title={Champ: Controllable and consistent human image animation with 3d parametric guidance},
  author={Zhu, Shenhao and Chen, Junming Leo and Dai, Zuozhuo and Dong, Zilong and Xu, Yinghui and Cao, Xun and Yao, Yao and Zhu, Hao and Zhu, Siyu},
  booktitle={European Conference on Computer Vision},
  pages={145--162},
  year={2024},
  organization={Springer}
}

@article{animateanyone2,
  title={Animate anyone 2: High-fidelity character image animation with environment affordance},
  author={Hu, Li and Wang, Guangyuan and Shen, Zhen and Gao, Xin and Meng, Dechao and Zhuo, Lian and Zhang, Peng and Zhang, Bang and Bo, Liefeng},
  journal={arXiv preprint arXiv:2502.06145},
  year={2025}
}

@ARTICLE{taghipour_video,
  author={Taghipour, Ashkan and Ghahremani, Morteza and Bennamoun, Mohammed and Miri Rekavandi, Aref and Li, Zinuo and Laga, Hamid and Boussaid, Farid},
  journal={IEEE Access}, 
  title={Faster Image2Video Generation: A Closer Look at CLIP Image Embedding’s Impact on Spatio-Temporal Cross-Attentions}, 
  year={2025},
  volume={13},
  number={},
  pages={141313-141327},
  doi={10.1109/ACCESS.2025.3595822}}

@ARTICLE{TIP_1,
  author={Wang, Zhaoyang and Hu, Bo and Zhang, Mingyang and Li, Jie and Li, Leida and Gong, Maoguo and Gao, Xinbo},
  journal={IEEE Transactions on Image Processing}, 
  title={Diffusion Model-Based Visual Compensation Guidance and Visual Difference Analysis for No-Reference Image Quality Assessment}, 
  year={2025},
  volume={34},
  number={},
  pages={263-278},
  doi={10.1109/TIP.2024.3523800}}

@ARTICLE{TIP_2,
  author={Fan, Guodong and Zhou, Yu and Zhou, Jingchun and Ju, Yakun and Chen, Guang-Yong and Li, Jinjiang and Kot, Alex C.},
  journal={IEEE Transactions on Image Processing}, 
  title={DCD-UIE: Decoupled Chromatic Diffusion Model for Underwater Image Enhancement}, 
  year={2026},
  volume={},
  number={},
  pages={1-1},
  doi={10.1109/TIP.2025.3648875}}

@ARTICLE{TIP_3,
  author={Bordin, Tom and Maugey, Thomas},
  journal={IEEE Transactions on Image Processing}, 
  title={Linearly Transformed Color Guide for Low-Bitrate Diffusion-Based Image Compression}, 
  year={2025},
  volume={34},
  number={},
  pages={468-482},
  doi={10.1109/TIP.2024.3521301}}

@article{ltx,
  title={Ltx-video: Realtime video latent diffusion},
  author={HaCohen, Yoav and Chiprut, Nisan and Brazowski, Benny and Shalem, Daniel and Moshe, Dudu and Richardson, Eitan and Levin, Eran and Shiran, Guy and Zabari, Nir and Gordon, Ori and others},
  journal={arXiv preprint arXiv:2501.00103},
  year={2024}
}

@article{svd,
  title={Stable Video Diffusion: Scaling Latent Video Diffusion Models to Large Datasets},
  author={Blattmann, Andreas and Dockhorn, Tim and Kulal, Sumith and Mendelevitch, Daniel and Kilian, Maciej and Lorenz, Dominik and others},
  journal={arXiv preprint arXiv:2311.15127},
  year={2023}
}

@article{human_video_survey,
  title={A Comprehensive Survey on Human Video Generation: Challenges, Methods, and Insights},
  author={Lei, Wentao and Wang, Jinting and Ma, Fengji and Huang, Guanjie and Liu, Li},
  journal={arXiv preprint arXiv:2407.08428},
  year={2024}
}

@article{animateanyone,
  title={Animate Anyone: Consistent and Controllable Image-to-Video Synthesis for Character Animation},
  author={Hu, Li and Gao, Xin and Zhang, Peng and Sun, Ke and Zhang, Bang and Bo, Liefeng},
  journal={arXiv preprint arXiv:2311.17117},
  year={2023}
}

@article{magicanimate,
  title={MagicAnimate: Temporally Consistent Human Image Animation using Diffusion Model},
  author={Xu, Zhongcong and Zhang, Jianfeng and Liew, Jun Hao and Yan, Hanshu and Liu, Jia-Wei and Zhang, Chenxu and Feng, Jiashi and Shou, Mike Zheng},
  journal={arXiv preprint arXiv:2311.16498},
  year={2023}
}

@article{dreamoving,
  title={DreaMoving: A Human Video Generation Framework based on Diffusion Models},
  author={Feng, Mengyang and Liu, Jinlin and Yu, Kai and Yao, Yuan and Hui, Zheng and Guo, Xiefan and Lin, Xianhui and Xue, Haolan and Shi, Chen and Li, Xiaowen and Li, Aojie and Kang, Xiaoyang and Lei, Biwen and Cui, Miaomiao and Ren, Peiran and Xie, Xuansong},
  journal={arXiv preprint arXiv:2312.05107},
  year={2023}
}

@article{dreampose,
  title={DreamPose: Fashion Image-to-Video Synthesis via Stable Diffusion},
  author={Karras, Johanna and Holynski, Aleksander and Wang, Ting-Chun and Kemelmacher-Shlizerman, Ira},
  journal={arXiv preprint arXiv:2304.06025},
  year={2023}
}

@article{simeoni2025dinov3,
  title={Dinov3},
  author={Sim{\'e}oni, Oriane and Vo, Huy V and Seitzer, Maximilian and Baldassarre, Federico and Oquab, Maxime and Jose, Cijo and Khalidov, Vasil and Szafraniec, Marc and Yi, Seungeun and Ramamonjisoa, Micha{\"e}l and others},
  journal={arXiv preprint arXiv:2508.10104},
  year={2025}
}

@ARTICLE{taghipur_b2b,
  author={Taghipour, Ashkan and Ghahremani, Morteza and Bennamoun, Mohammed and Rekavandi, Aref Miri and Laga, Hamid and Boussaid, Farid},
  journal={IEEE Transactions on Multimedia}, 
  title={Box It to Bind It: Unified Layout Control and Attribute Binding in Text-to-Image Diffusion Models}, 
  year={2025},
  volume={27},
  number={},
  pages={8393-8407},
  doi={10.1109/TMM.2025.3607759}}

@article{skyreels,
  title={Skyreels-a1: Expressive portrait animation in video diffusion transformers},
  author={Qiu, Di and Fei, Zhengcong and Wang, Rui and Bai, Jialin and Yu, Changqian and Fan, Mingyuan and Chen, Guibin and Wen, Xiang},
  journal={arXiv preprint arXiv:2502.10841},
  year={2025}
}

@inproceedings{clearclip,
  title={ClearCLIP: Decomposing CLIP Representations for Dense Vision-Language Inference},
  author={Lan, Mengcheng and Zhong, Chaofeng and Jiang, Yiping and Ke, Qingyan and Gu, Yu and Zhao, Qiong and Zou, Wenqi},
  booktitle={European Conference on Computer Vision (ECCV)},
  pages={1--17},
  year={2024}
}

@inproceedings{cloc,
  title={Contrastive Localized Language-Image Pre-Training},
  author={Chen, Hong-You and Zhou, Zhengfeng and Shang, Haotian and Pouransari, Hadi and Wu, Yifei and Fang, Alex and Harber, Evan and Lim, Ser-Nam and Jayasuriya, Suren and Tuzel, Oncel},
  booktitle={Proceedings of the IEEE/CVF Conference on Computer Vision and Pattern Recognition (CVPR)},
  pages={14212--14222},
  year={2024}
}

@inproceedings{clip2dino,
  title={From CLIP to DINO: Visual Encoders Shout in Multi-modal Large Language Models},
  author={Wu, Dongsheng and Liang, Chaoyou and Chen, Xiaoyi and Gao, Zhiqi and Han, Songxin and Li, Yuqing and Li, Kunpeng and Wang, Fan and Yan, Rui and Xie, Xuguang},
  booktitle={International Conference on Learning Representations (ICLR)},
  year={2024}
}

@inproceedings{consisid,
  title={Identity-Preserving Text-to-Video Generation by Frequency Decomposition},
  author={Yuan, Shenghai and Huang, Jinfa and He, Xianyi and Ge, Yunyuan and Shi, Yujun and Chen, Liuhan and Luo, Jiebo and Yuan, Li},
  booktitle={Proceedings of the IEEE/CVF Conference on Computer Vision and Pattern Recognition (CVPR)},
  pages={8218--8228},
  year={2025}
}

@article{mofe,
  title={From Large Angles to Consistent Faces: Identity-Preserving Video Generation via Mixture of Facial Experts},
  author={Wang, Zhenzhong and Zhang, Junhao and Hu, Jia and Zhang, Jianfeng and Zhang, Liang and Liu, Ziwei},
  journal={arXiv preprint arXiv:2508.09476},
  year={2025}
}

@inproceedings{disco,
  title={DisCo: Disentangled Control for Realistic Human Dance Generation},
  author={Wang, Tan and Li, Linjie and Lin, Kevin and Lin, Chung-Ching and Yang, Zhengyuan and Liu, Zicheng and Wang, Lijuan},
  booktitle={Proceedings of the IEEE/CVF Conference on Computer Vision and Pattern Recognition (CVPR)},
  pages={9326--9336},
  year={2024}
}

@article{magicpose,
  title={MagicPose: Realistic Human Poses and Facial Expressions Retargeting with Identity-aware Diffusion},
  author={Chang, Di and Shi, Yichun and Gao, Quankai and Fu, Jessica and Xu, Hongyi and Song, Guoxian and Yan, Qing and Yang, Xiao and Soleymani, Mohammad},
  journal={arXiv preprint arXiv:2311.12052},
  year={2023}
}

@article{animate_x,
  title={Animate-X: Universal Character Image Animation with Enhanced Motion Representation},
  author={Tan, Shuai and Gu, Biao and Wu, Yingqing and Zhang, Linchao and Wu, Jiahao and Gao, Fei and Yan, Jianfei and Liu, Xiao},
  journal={arXiv preprint arXiv:2411.10170},
  year={2024}
}

@article{stableanimator,
  title={StableAnimator: High-Quality Identity-Preserving Human Image Animation},
  author={Tu, Shuyuan and Liao, Zhen and Zhao, Xujie and Huang, Zhaoyang and Xu, Ziwei and Liu, Yuan and Liu, Ziwei},
  journal={arXiv preprint arXiv:2411.17697},
  year={2024}
}

@article{human4dit,
  title={Human4DiT: Free-view Human Video Generation with 4D Diffusion Transformer},
  author={Shao, Ruizhi and Zhang, Youxin and Liu, Zhengming and Sun, Hongwen and Han, Yuxiang and Wang, Wenping and Zhou, Xiaowei},
  journal={arXiv preprint arXiv:2405.17405},
  year={2024}
}

@inproceedings{followyourpose,
  title={Follow Your Pose: Pose-Guided Text-to-Video Generation using Pose-Free Videos},
  author={Ma, Yue and He, Yingqing and Cun, Xiaodong and Wang, Xintao and Chen, Siran and Li, Xiu and Shan, Ying},
  booktitle={Proceedings of the AAAI Conference on Artificial Intelligence},
  volume={38},
  pages={4117--4125},
  year={2024}
}

@article{controlnext,
  title={ControlNeXt: Powerful and Efficient Control for Image and Video Generation},
  author={Peng, Bohao and Wang, Jian and Zhang, Yuechen and Li, Wenbo and Yang, Ming-Chang and Jia, Jiaya},
  journal={arXiv preprint arXiv:2408.06070},
  year={2024}
}

@article{mosa,
  title={Mosa: Motion generation with scalable autoregressive modeling},
  author={Liu, Mengyuan and Yan, Sheng and Wang, Yong and Li, Yingjie and Bian, Gui-Bin and Liu, Hong},
  journal={arXiv preprint arXiv:2511.01200},
  year={2025}
}

@inproceedings{vbench,
  title={Vbench: Comprehensive benchmark suite for video generative models},
  author={Huang, Ziqi and He, Yinan and Yu, Jiashuo and Zhang, Fan and Si, Chenyang and Jiang, Yuming and Zhang, Yuanhan and Wu, Tianxing and Jin, Qingyang and Chanpaisit, Nattapol and others},
  booktitle={Proceedings of the IEEE/CVF Conference on Computer Vision and Pattern Recognition},
  pages={21807--21818},
  year={2024}
}

@article{lpips,
  title={The Unreasonable Effectiveness of Deep Features as a Perceptual Metric},
  author={Richard Zhang and Phillip Isola and Alexei A. Efros and Eli Shechtman and Oliver Wang},
  journal={2018 IEEE/CVF Conference on Computer Vision and Pattern Recognition},
  year={2018},
  pages={586-595},
  url={https://api.semanticscholar.org/CorpusID:4766599}
}

@article{FVD,
  title={Towards accurate generative models of video: A new metric \& challenges},
  author={Unterthiner, Thomas and Van Steenkiste, Sjoerd and Kurach, Karol and Marinier, Raphael and Michalski, Marcin and Gelly, Sylvain},
  journal={arXiv preprint arXiv:1812.01717},
  year={2018}
}

@ARTICLE{SSIM,
  author={Zhou Wang and Bovik, A.C. and Sheikh, H.R. and Simoncelli, E.P.},
  journal={IEEE Transactions on Image Processing}, 
  title={Image quality assessment: from error visibility to structural similarity}, 
  year={2004},
  volume={13},
  number={4},
  pages={600-612},
  doi={10.1109/TIP.2003.819861}}

@INPROCEEDINGS{psnr,
  author={Horé, Alain and Ziou, Djemel},
  booktitle={2010 20th International Conference on Pattern Recognition}, 
  title={Image Quality Metrics: PSNR vs. SSIM}, 
  year={2010},
  volume={},
  number={},
  pages={2366-2369},
  doi={10.1109/ICPR.2010.579}}

@inproceedings{raft,
  title={Raft: Recurrent all-pairs field transforms for optical flow},
  author={Teed, Zachary and Deng, Jia},
  booktitle={European conference on computer vision},
  pages={402--419},
  year={2020},
  organization={Springer}
}

@inproceedings{musiq,
  title={Musiq: Multi-scale image quality transformer},
  author={Ke, Junjie and Wang, Qifei and Wang, Yilin and Milanfar, Peyman and Yang, Feng},
  booktitle={Proceedings of the IEEE/CVF international conference on computer vision},
  pages={5148--5157},
  year={2021}
}

@inproceedings{ntu_dataset,
  title={Ntu rgb+ d: A large scale dataset for 3d human activity analysis},
  author={Shahroudy, Amir and Liu, Jun and Ng, Tian-Tsong and Wang, Gang},
  booktitle={Proceedings of the IEEE conference on computer vision and pattern recognition},
  pages={1010--1019},
  year={2016}
}

@INPROCEEDINGS{kth_dataset,
  author={Schuldt, C. and Laptev, I. and Caputo, B.},
  booktitle={Proceedings of the 17th International Conference on Pattern Recognition, 2004. ICPR 2004.}, 
  title={Recognizing human actions: a local SVM approach}, 
  year={2004},
  volume={3},
  number={},
  pages={32-36 Vol.3},
  doi={10.1109/ICPR.2004.1334462}}

@INPROCEEDINGS{hmdb_dataset,
  author={Kuehne, H. and Jhuang, H. and Garrote, E. and Poggio, T. and Serre, T.},
  booktitle={2011 International Conference on Computer Vision}, 
  title={HMDB: A large video database for human motion recognition}, 
  year={2011},
  volume={},
  number={},
  pages={2556-2563},
  doi={10.1109/ICCV.2011.6126543}}

@article{ucf_dataset,
  title={Ucf101: A dataset of 101 human actions classes from videos in the wild},
  author={Soomro, Khurram and Zamir, Amir Roshan and Shah, Mubarak},
  journal={arXiv preprint arXiv:1212.0402},
  year={2012}
}


% \end{IEEEbiography}


%\newpage



%\vfill


\end{document}